\documentclass[graybox]{svmult} %vecphys italicizes vectors

\usepackage{mathptmx}       % selects Times Roman as basic font
\usepackage{helvet}         % selects Helvetica as sans-serif font
\usepackage{courier}        % selects Courier as typewriter font
\usepackage{type1cm}        % activate if the above 3 fonts are
\usepackage{makeidx}         % allows index generation
\usepackage{graphicx}        % standard LaTeX graphics tool
\usepackage{multicol}        % used for the two-column index
\usepackage[bottom]{footmisc}% places footnotes at page bottom

\makeindex             % used for the subject index
\usepackage{geometry}
\usepackage{amsmath} %amsmath is necessary, amssymb,amsfonts aren't
\usepackage{cite}  %cite - increasing order of references; ",enumerate" - ?
\usepackage{bm}

\usepackage{floatrow} %used in {H}
\usepackage{booktabs}   %used
\usepackage{multirow}   %used \multirow

\def\Up#1{\vspace{-#1em}}
\def\x#1{} %to mark corrections; \x{??} \x{!!}
\def\diag{\operatorname{diag}}
\def\Diag{\operatorname{Diag}}
\def\vol{\operatorname{vol}}
\def\b{t} %\beta (former)
\def\e{\vec e}
\def\ul#1{\underline{#1}}

\def\xy{\hspace{.07em}}

      \title*{Do logarithmic proximity measures outperform plain ones in graph %nodes\x{+}
              clustering?}
\titlerunning{Do logarithmic proximity measures outperform plain ones in graph %nodes\x{+}
              clustering?}
\author{Vladimir Ivashkin and Pavel Chebotarev\thanks{The work of this author was supported by the Russian Science Foundation [grant 16-11-00063].}}%\\
\institute{Vladimir Ivashkin \at Moscow Institute of Physics and Technology, 9 Inststitutskii per., Dolgoprudny, Moscow region, 141700 Russia, \email{vladimir.ivashkin@phystech.edu}
\and Pavel Chebotarev \at Institute of Control Sciences of the Russian Academy of Sciences, 65 Profsoyuznaya str., Moscow, 117997 Russia\,
and Moscow Institute of Physics and Technology, 9 Inststitutskii per., Dolgoprudny, Moscow region, 141700 Russia, \email{pavel4e@gmail.com}}
\date{}
\authorrunning{V.~Ivashkin and P.~Chebotarev}
\begin{document}

\maketitle
\setcounter{footnote}{0}

\Up{4.2}
\abstract{We consider a number of graph kernels and proximity measures including commute time kernel, regularized Laplacian kernel, heat kernel, exponential diffusion kernel (also called ``communicability''), etc., and the corresponding distances as applied to clustering nodes in random graphs and several well-known datasets. The model of generating random graphs involves edge probabilities for the pairs of nodes that belong to the same class or different predefined classes of nodes. It turns out that in most cases, logarithmic measures (i.e., measures resulting after taking logarithm of the proximities) perform better while distinguishing underlying classes than the ``plain'' measures. A comparison in terms of reject curves of inter-class and intra-class distances confirms this conclusion. A similar conclusion can be made for several well-known datasets. A possible origin of this effect is that most kernels have a multiplicative nature, while the nature of distances used in cluster algorithms is an additive one (cf. the triangle inequality). The logarithmic transformation is a tool to transform the first nature to the second one. Moreover, some distances corresponding to the logarithmic measures possess a meaningful cutpoint additivity property. In our experiments, the leader is usually the logarithmic Communicability measure. However, we indicate some more complicated cases in which other measures, typically, Communicability and plain Walk, can be the winners.}

\section{Introduction}
\label{s_intro}

In this paper, we consider a number of graph kernels and proximity measures and the corresponding distances as applied to clustering nodes in random graphs and several %well-known
datasets. The measures include the commute time kernel, the regularized Laplacian kernel, the heat kernel, the exponential diffusion kernel, and some others. The model $G(N, (m)p_\text{in}, p_\text{out})$ of generating random graphs involves edge probabilities for the pairs of nodes that belong to the same class ($p_\text{in}$) or different classes ($p_\text{out}$). For a review on graph clustering we refer the reader to~\cite{Schaeffer07CSR,Fortunato10FR,FoussSaerensShimbo16}.

The main result of the present study is that in a number of simple cases, logarithmic measures (i.e., measures resulting after taking logarithm of the proximities) perform better while distinguishing underlying classes than the ``plain'' measures. A direct comparison, in terms of ROC curves, of inter-class and intra-class distances confirms this conclusion. However, there are exceptions to that rule.
In most experiments, the leader is the new measure logComm (logarithmic Communicability).

%\section{Measures}
%\label{s_measures}

Recall that if a proximity measure satisfies the \emph{triangle inequality for proximities\/} $p(x,y)+p(x,z)-p(y,z)\le p(x,x)$ for all nodes $x,y,z\in V(G),$ then the function $d(x,y)= p(x,x)+p(y,y)-p(x,y)-p(y,x)$ satisfies the ordinary triangle inequality~\cite{CheSha98a}. In this study, we constantly rely on the duality between metrics and proximity measures.

The paper is organized as follows. In the remainder of Section~\ref{s_intro}, we present the metrics and proximity measures under study.
In Section~\ref{s_logVS}, the logarithmic and plain measures are juxtaposed on several clustering tasks with random graphs generated by the $G(N, (m)p_\text{in}, p_\text{out})$ model with a small number of classes~$m.$
In Section~\ref{s_compet}, 13 measure families compete in two tournaments generated by eight clustering tasks with different parameters. The first tournament gathers the best representatives of each family; the participants of the second one are the representatives with suboptimal parameters corresponding to the 90th percentiles. Every task involves the generation of 50 random graphs.
Section~\ref{s_reject} presents a different way of comparing the proximity measures: it is based on drawing the ROC curves. This kind of comparison only deals with inter-class and intra-class distances and does not depend on the specific clustering algorithm.
In Section~\ref{s_diffsize}, we extend the set of tests: here, the classes of nodes have different sizes, while the intra-class and inter-class edge probabilities are not uniform. Finally, in Section~\ref{s_classic}, from random graphs we turn to several classical datasets and make the measure families to meet in two new tournaments. In the concluding Section~\ref{s_Conclu}, we briefly  discuss the results.

%In this study, we consider the graph measures listed below.
Thus, in the following subsections, we list the families of node proximity measures \cite{CheSha98}, including kernels\footnote{On various graph kernels, see, e.g., \cite{FoussFrancoisseSaerens11}.}, and distances, which have been proposed in the literature and have proven to be practical. Generally speaking, our main goal is to find the measures that are the most practical.

%\end{document}
\subsection{The Shortest path and Commute time distances}

\begin{itemize}
	\item The \textbf{Shortest Path} distance $d^s(i, j)$ on a graph $G=(V,E)$ is the length of a shortest path between $i$ and $j$ in $G$~\cite{BuckleyHarary90}.
	\item The \textbf{Commute Time} distance $d^c(i, j)$ is the average length of random walks from $i$ to $j$ and back. The transition probabilities of the corresponding Markov chain are obtained by normalizing the rows of the adjacency matrix of~$G.$ This distance is related to the Commute-time kernel \cite{Saerens2004principal} $K_\text{CT}=L^+,$ the pseudoinverse of the Laplacian matrix $L$ of~$G.$
	\item The \textbf{Resistance} distance \cite{Sharpe67a,GvishianiGurvich87En,KleinRandic93} $d^r(i, j)$ is the effective resistance between $i$ and $j$ in the resistive electrical network corresponding to~$G.$

The Resistance distance is well known \cite{Nash-Williams59,GobelJagers74RandomW,ChandraRaghavanRuzzoSmolenskyTiwari89} to be proportional to the Commute Time distance. %in the corresponding Markov chain.

Let $D^s$ and $D^r$ be the matrices of shortest path distances and resistance distances in $G,$ respectively. As we mainly study parametric families of graph measures, for comparability, the parametric family $(1-\lambda)D^s + \lambda D^r$ with $\lambda\in[0,\,1]$ (i.e., the convex combination of the \textbf{Shortest Path} distance and the \textbf{Resistance} distance) will be considered. We will denote it by \textbf{SP--CT}.
\end{itemize}

\subsection{The plain Walk, Forest, Communicability, and Heat kernels / proximities}
Now we introduce the short names of node proximity measures related to several families of graph kernels.
\begin{itemize}
	\item \textbf{plain Walk} (Von Neumann diffusion kernel)
		$K_t^{\text{pWalk}} = (I - tA)^{-1},\: 0 < t < \rho^{-1}$ ($\rho$ is the spectral radius of $A,$ the adjacency matrix of~$G$) \cite{CheSha98,Kandola2002learning}.
	\item \textbf{Forest} (Regularized Laplacian kernel):
		$K_t^{\text{For}} = (I + tL)^{-1},\: t > 0,$ where $L$ is the Laplacian matrix of $G$ \cite{CheSha97,CheSha01,SmolaKondor03}.
	\item \textbf{Communicability} (Exponential diffusion kernel):
		$K_t^{\text{Comm}} = \exp(tA),\: t > 0$ \cite{KondorLafferty02diffusion,Estrada12LAA}.
	\item \textbf{Heat kernel} (Laplacian exponential diffusion kernel):
		$K_t^{\text{Heat}} = \exp(-tL),\: t > 0$ \cite{ChungYau98,KondorLafferty02diffusion}.
\end{itemize}

\subsection{Logarithmic measures \emph{\cite{Che11AAM}:} Walk, Forest, Communicability, and Heat}
\begin{itemize}
	\item \textbf{Walk (logarithmic):}
		$K_t^{\text{Walk}} = \overrightarrow{\ln K_t^{\text{pWalk}}},\: 0 < t < \rho^{-1}$, where $\overrightarrow{\ln K}$ is the element-wise $\ln$ of a matrix $K$~\cite{Che12DAM}.
	\item \textbf{logarithmic Forest:}
		$K_t^{\text{logFor}} = \overrightarrow{\ln K_t^{\text{For}}},\: t > 0$ \cite{Che11DAM}.
	\item \textbf{logarithmic Communicability:}
		$K_t^{\text{logComm}} = \overrightarrow{\ln K_t^{\text{Comm}}},\: t > 0.$
	\item \textbf{logarithmic Heat:}
		$K_t^{\text{logHeat}} = \overrightarrow{\ln K_t^{\text{Heat}}},\: t > 0.$
\end{itemize}

\subsection{Sigmoid Commute Time and Sigmoid Corrected Commute Time kernels \emph{\cite{LuxburgRadlHein09,SommerFoussSaerens16ICANN,FoussSaerensShimbo16}}}
The Corrected Commute Time kernel is defined by
$$
	K_\text{CCT} = HD^{-\frac12}M(I-M)^{-1}MD^{-\frac12}H\quad\mbox{with}\quad M=D^{-\frac12}\Big(A-\frac{\vec d\vec d^T}{\vol(G)}\Big)D^{-\frac12},
$$
where $H = I - \e\e^T/N$ is the centering matrix, %that removes in each column of the distance matrix the mean of the column,
$\e=(\underbrace{1,\ldots,1}_N)^T,\,$ $\vec d=A\e,\,$ $D = \diag(\vec d),\,$ $\diag(\vec v)$ is the diagonal matrix with vector $\vec v$ on the diagonal, and $\vol(G)=|V|,\,$ $V$ being the edge set of~$G.$

Applying the element-wise sigmoid transformation to $K_\text{CT}$ and $K_\text{CCT}$ we obtain the corresponding \emph{sigmoid kernels\/} $K^S$:
$$
	[K^S]_{ij} = \frac{1}{1 + \exp(-tk_{ij}/\sigma)},\quad i,j=1,\ldots,N,
$$
where $k_{ij}$ is an element of a kernel matrix ($K_\text{CT}$ or $K_\text{CCT}$), $t$ is a parameter, and $\sigma$ is the standard
deviation of the elements of the kernel matrix. The \textbf{Sigmoid Commute Time kernel} and \textbf{Sigmoid Corrected Commute Time kernel} will be abbreviated as \textbf{SCT} and \textbf{SCCT}, respectively.

\subsection{Randomized Shortest Path and Free Energy dissimilarity measures \emph{\cite{Kivimaki2014developments}}}
\label{s_FE}

\begin{itemize}
	\item \textbf{Preliminaries}:\\
		$$
			P^{\,\text{ref}} = D^{-1}A,\quad\text{where } D = \diag(A\e); %,\quad e=(\underbrace{1,\ldots,1}_N)^T;
		$$
		$$
			W = P^{\,\text{ref}} \circ \overrightarrow{\exp(-\b C)},\quad \text{where } \circ \text{ is element-wise product},
		$$
		$C$ is the matrix of the Shortest Path distances, $\b$ being the ``inverse temperature'' parameter;\\
		$$
			Z = (I - W)^{-1}.
		$$
	\item \textbf{Randomized Shortest Path (RSP)}:\\
		$$
			S = (Z(C \circ W)Z) \div Z,\quad  \text{ where } \div \text{ is element-wise division};
		$$
		$$
		\bar C = S - \e(\vec d_S)^T;\quad \vec d_S = \diag(S),\;\text{ where } \diag(S) \text{ is the vector on the diagonal of square matrix }S;
		$$
		$$
			\Delta_\text{\,RSP} = (\bar{C} + \bar{C}^{\xy T})/2.
		$$
	\item \textbf{Helmholtz Free Energy distance (FE)}:\\
		$$
			Z^{\xy h} = Z D_h^{-1},\quad\text{ where } D_h = \Diag(Z),
		$$
		where $\Diag(Z)$ denotes the diagonal matrix whose diagonal coincides with that of~$Z;$
		$$
			\Phi = -\b^{-1}\,\overrightarrow{\ln Z^{\xy h}s};
		$$
		$$
			\Delta_\text{\,FE} = (\Phi + \Phi^T)/2.
		$$		
\end{itemize}

%\vspace*{-2\baselineskip}
As we know from the classical scaling theory, the inner product matrix (which is a kernel) can be obtained from a [Euclidean] distance matrix $\Delta$ by the
$$%\begin{equation}
	K = -\frac{1}{2}H\Delta^{(2)}H
%	\label{eq:kernel}
$$%\end{equation}
transformation, where $H = I - \e\e^T/N\,$ is the centering matrix. %that removes in each column of the distance matrix the mean of the column.

\medskip
For comparability, all family parameters are adjusted to the $[0,\,1]$ segment by a linear transformation or some $t/(t+c)$ transformation or both of them.

The comparative behavior of graph kernels in clustering tasks has been studied in several papers, including \cite{Kivimaki2014developments,ColletteA15Master,SommerFoussSaerens16ICANN}. The originality of our approach is that (1) we do not fix the family parameters and rather optimize them during the experiments, (2)~we compare a larger set of measure families, and (3) we juxtapose logarithmic and plain measures.

\section{Logarithmic vs. plain measures}
\label{s_logVS}

Let $G(N, (m)p_\text{in}, p_\text{out})$ be the model of generating random graphs on $N$ nodes divided into $m$ classes of the same size, with $p_\text{in}$ and $p_\text{out}$ being the probability of $(i,j)\in E(G)$ for $i$ and $j$ belonging to the same class and different classes, respectively, where $E(G)$ is the edge set of~$G.$

\begin{figure}[H]
%\samenumber%\subfigures
	\begin{minipage}{.245\textwidth}
        \leftfigure{\includegraphics[width=.99\linewidth]{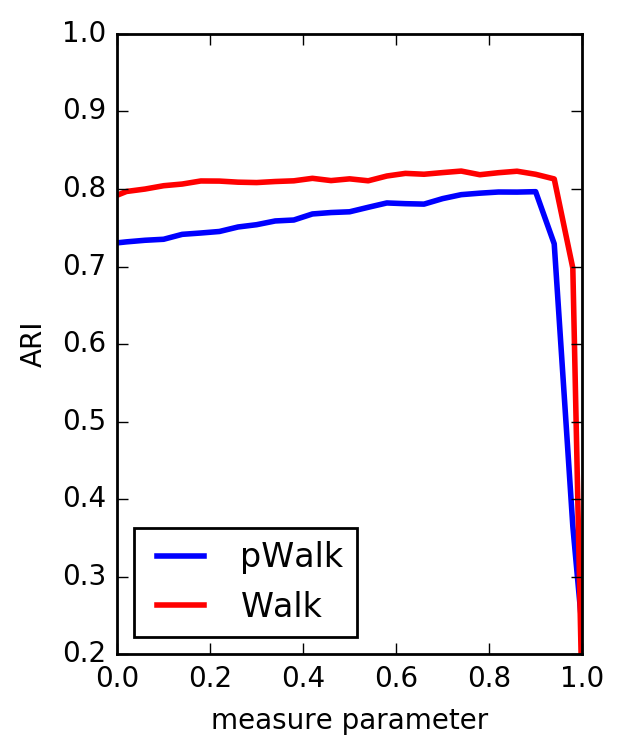}}
        \\\centerline{(a) plain Walk and Walk}
	\end{minipage}%\hspace{\fill}
	\begin{minipage}{.245\textwidth}
		\leftfigure{\includegraphics[width=.99\linewidth]{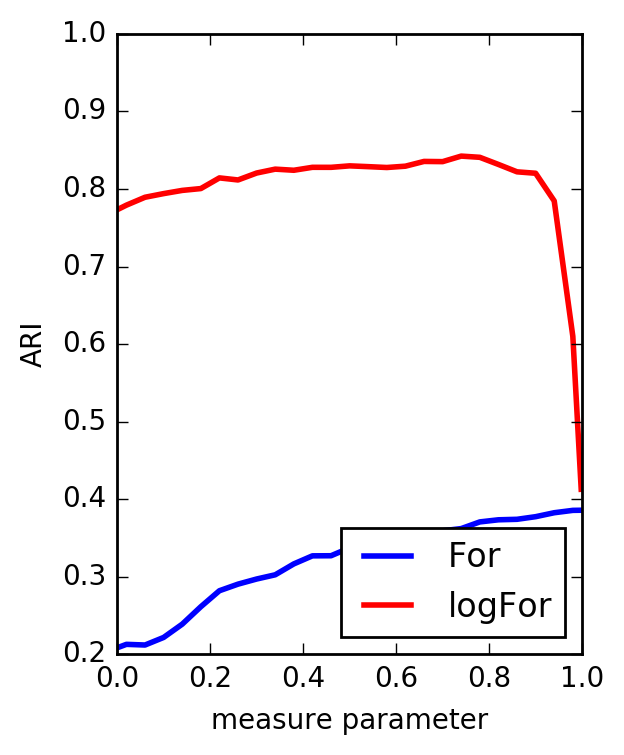}}
        \\\centerline{(b) Forest and logarithmic Forest}
	\end{minipage}
	\begin{minipage}{.245\textwidth}
		\leftfigure{\includegraphics[width=.99\linewidth]{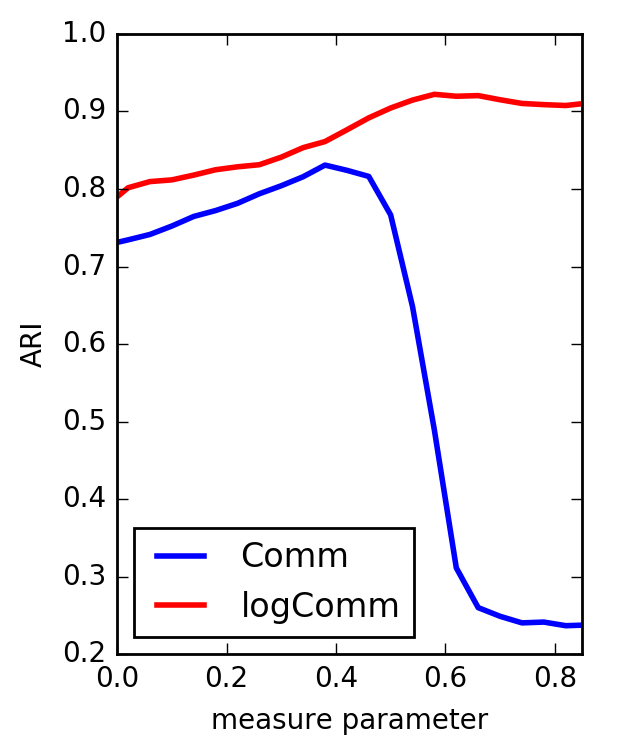}}
        \\\centerline{(c) (log)Communicability}
	\end{minipage}%
	\begin{minipage}{.245\textwidth}
		\leftfigure{\includegraphics[width=.99\linewidth]{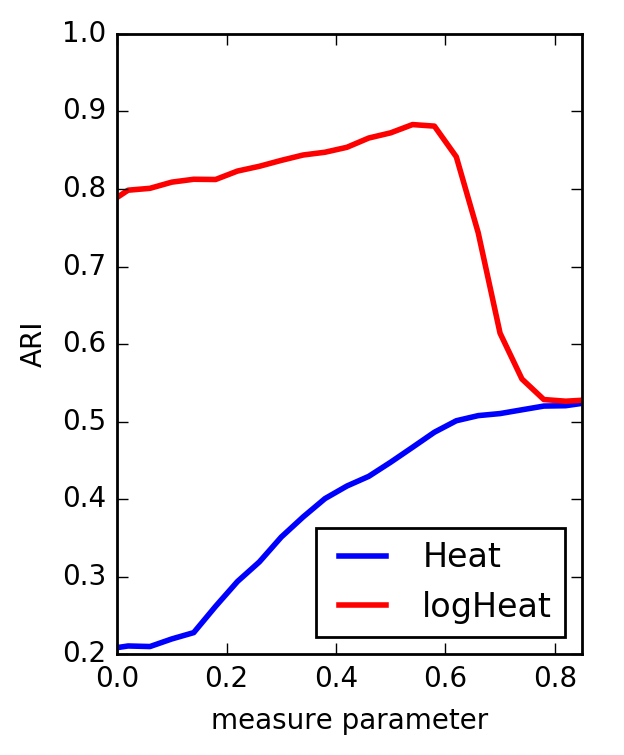}}
        \\\centerline{(d) Heat and logarithmic Heat}
	\end{minipage}
	\caption{\label{f_vs1}Logarithmic vs. plain measures for $G(100, (2)0.2, 0.05)$}
\end{figure}

\begin{figure}[H]
	\begin{minipage}{.245\textwidth}
		\leftfigure{\includegraphics[width=.99\linewidth]{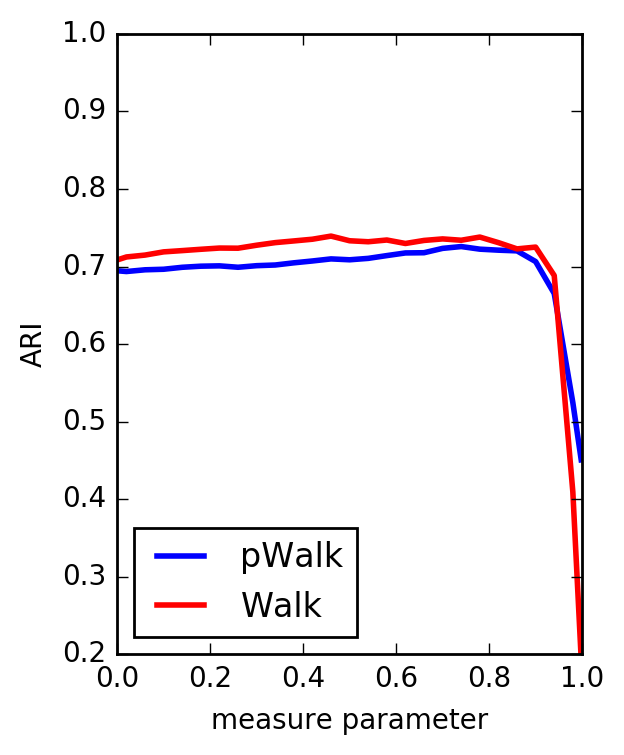}}
		\\\centerline{(a) plain Walk and Walk}
	\end{minipage}%
	\begin{minipage}{.245\textwidth}
		\leftfigure{\includegraphics[width=.99\linewidth]{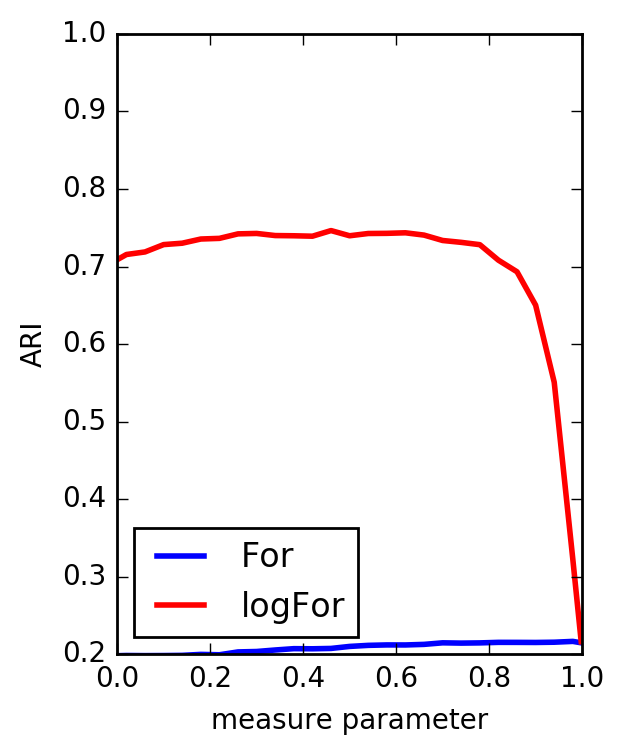}}
		\\\centerline{(b) Forest and logarithmic Forest}
	\end{minipage}
	\begin{minipage}{.245\textwidth}
		\leftfigure{\includegraphics[width=.99\linewidth]{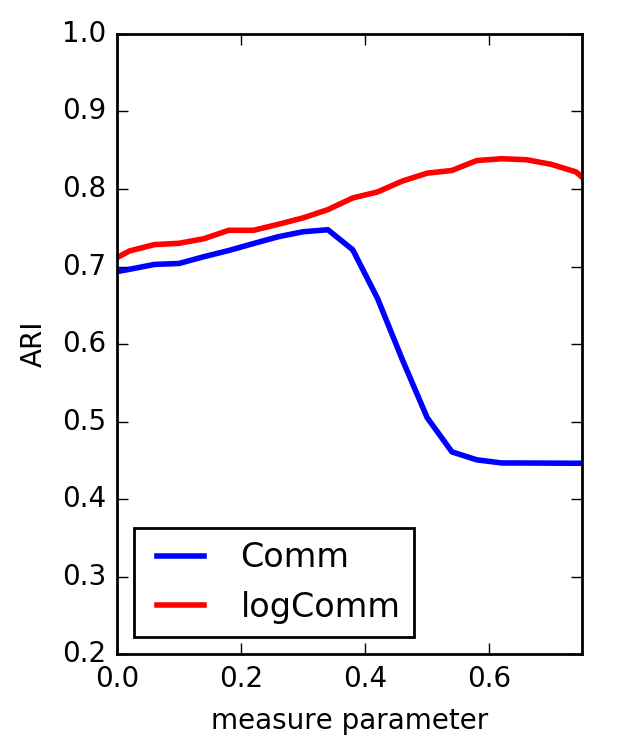}}
        \\\centerline{(c) (log)Communicability}
	\end{minipage}%
	\begin{minipage}{.245\textwidth}
		\leftfigure{\includegraphics[width=.99\linewidth]{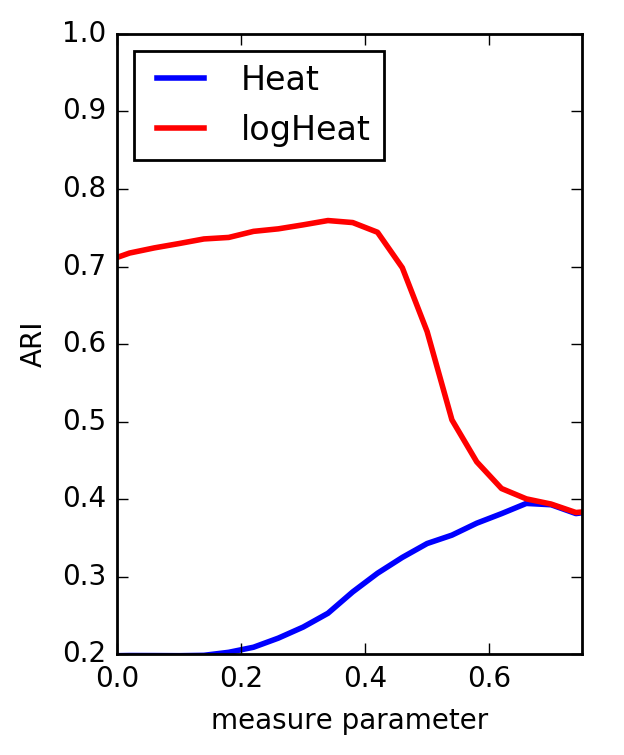}}
		\\\centerline{(d) Heat and logarithmic Heat}
	\end{minipage}
	\caption{\label{f_vs2}Logarithmic vs. plain measures for $G(100, (3)0.3, 0.1)$}
\end{figure}

\begin{figure}[H]
	\begin{minipage}{.245\textwidth}
		\leftfigure{\includegraphics[width=.99\linewidth]{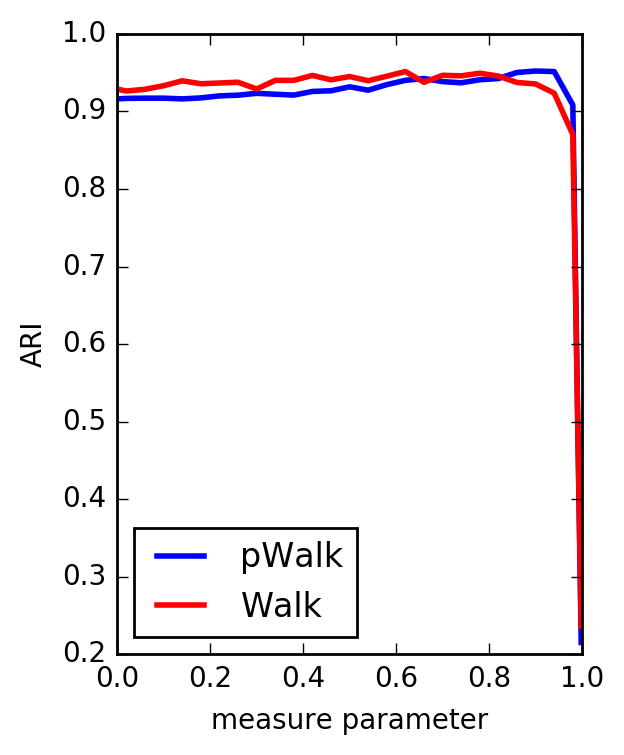}}
		\\\centerline{(a) plain Walk and Walk}
	\end{minipage}%
	\begin{minipage}{.245\textwidth}
		\leftfigure{\includegraphics[width=.99\linewidth]{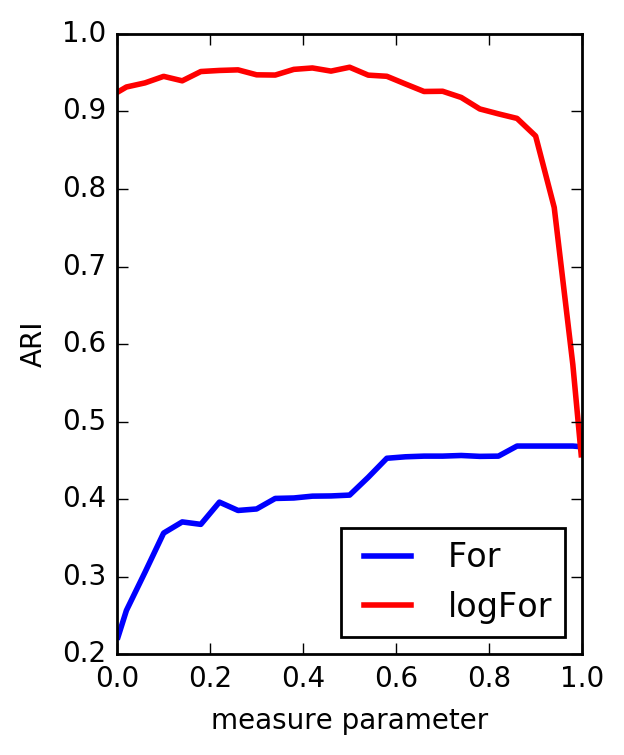}}
		\\\centerline{(b) Forest and logarithmic Forest}
	\end{minipage}
	\begin{minipage}{.245\textwidth}
		\leftfigure{\includegraphics[width=.99\linewidth]{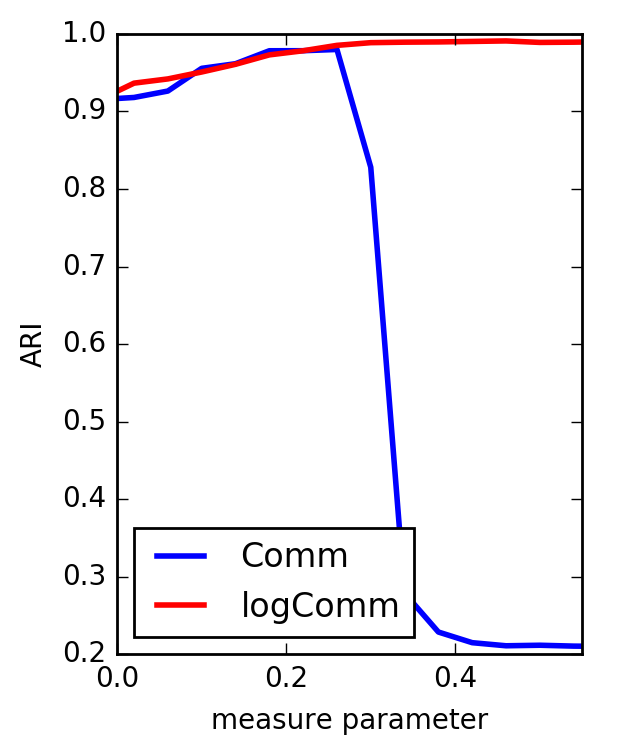}}
        \\\centerline{(c) (log)Communicability}
	\end{minipage}%
	\begin{minipage}{.245\textwidth}
		\leftfigure{\includegraphics[width=.99\linewidth]{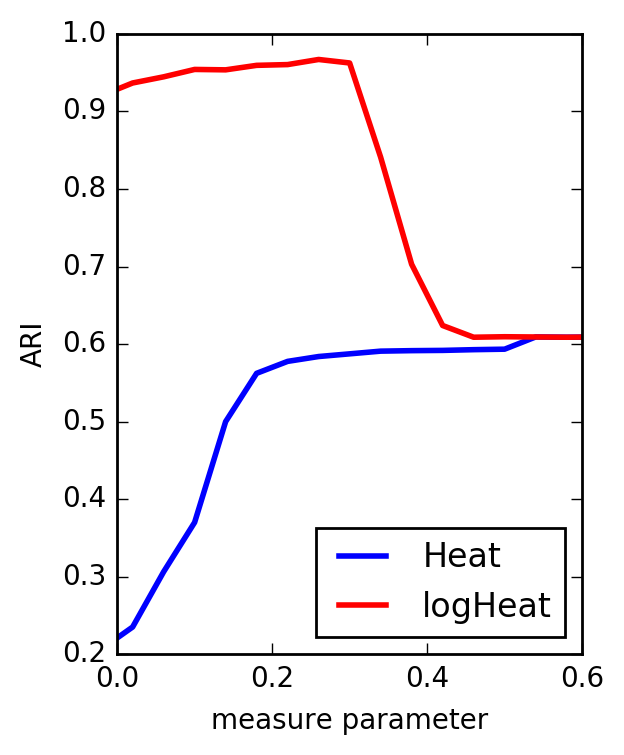}}
		\\\centerline{(d) Heat and logarithmic Heat}
	\end{minipage}
	\caption{\label{f_vs3}Logarithmic vs. plain measures for $G(200, (2)0.3, 0.1)$}
\end{figure}

The curves in Figures~\ref{f_vs1}--\ref{f_vs3} present the adjusted Rand index\footnote{On \emph{Rand index\/} (RI) and \emph{adjusted Rand index\/} (ARI) we refer to~\cite{HubertArabie85}.} (averaged over 200 random graphs) for clustering with Ward's method~\cite{Ward63JASA}.

\Up{2}
\nopagebreak

\indent It can be seen that in almost all cases, logarithmic measures outperform %dominate, majorize
the ordinary ones. The only exception, where the situation is ambiguous, is the case of Walk measures for random graphs on 200 nodes.

\section{Competition by Copeland's score}
\label{s_compet}

In this section, we present the results of many clustering tests in the form of tournaments whose participants are the measure families. Every family is characterized by its Copeland's score, i.e., the difference between the numbers of ``wins'' and ``losses'' in paired confrontations with the other families.

\subsection{Approach \emph{\cite{Kivimaki2014developments}}}
	\begin{itemize}
		\item The competition of measure families is based on paired comparisons.
		\item Every time when the best adjusted Rand index (ARI) of a measure family $F_1$ is higher on a random test graph than that of some other measure family $F_2$, we add $+1$ to the score of $F_1$ and $-1$ to the score of $F_2$.
	\end{itemize}

\Up{.8}
\subsection{The competition results}
The competition has been performed on random graphs generated with the $G(N, (m)p_\text{in}, p_\text{out})$ model and the following parameters: $N\!\in\!\{100, 200\},$ $\text{the number of classes }m\!\in\!\{2, 4\},$ $p_\text{in}=0.3,$ $p_\text{out}\!\in\!\{0.1, 0.15\}$. For every combination of parameters, we generated 50 graphs and for each of them we computed the best ARI's the measure families reached.
The results are presented in Table~\ref{t_CopComp}(a).
%The common parameters are: $p_\text{in} = 0.3$; 5 graphs for each competition.

\subsection{A competition for 90th percentiles} % Copeland's method: Ward,

%To check the robustness, the same competition has been done for the 90th percentiles:
Whenever we are looking for the best parameter of a measure family, we compute ARI on a grid of that parameter. In the above competition, we only compared the highest ARI values. Now consider the set of ARI values some measure family provides as a sample and find its 90th percentile. These percentiles become the participants in another tournament. The motivation behind this approach is to take into account the robustness of each family.
%So, if measure family is robust with respect to parameter, its 90th percentile of ARIs will be almost the same like the highest ARI, but otherwise its results will be significantly lower.

The results of the competition for 90th percentiles are given in Table~\ref{t_CopComp}(b).

\begin{table}[H]{\small
	\begin{minipage}{.49\textwidth}		
		\centering
		\begin{tabular}{lrrrrrrrrr}
			\toprule
			\multicolumn{1}{r}{\textbf{Nodes}}         & 100&    100& 100&  100& 200&  200& 200&  200&{\textbf{Sum}}\\
			\multicolumn{1}{r}{\textbf{Classes}}	   &   2&      2&   4&    4&   2&    2&   4&    4&{\textbf{  of}}\\  %\multicolumn{1}{l}
			\multicolumn{1}{r}{$\bm{p_{\mathbf{out}}}$}& 0.1&	0.15& 0.1& 0.15& 0.1& 0.15& 0.1& 0.15&{\textbf{scores}}\\
			\midrule
logComm\!\!&   404&	   539&	   453& 	391&	235&	578&	598&	590&	$\bm{3788}$\\
   SCCT&	   298&    299&    341& 	275&	297&	415&	454&	454&	$\bm{2833}$\\
 logFor&	   154&    182&    202& 	226&	207&	44&	    226&	192&	$\bm{1433}$\\
logHeat&	   249&    261&	   140&	     28&	175&	302&	251&  $-64$&	$\bm{1342}$\\
     FE&	    71&	    88&	   161&	    208&	 77&	 63& 	 82&	160&	$\bm{910}$\\
   Comm&	   120&	     9&	    27&	   $-2$&	267&	138&	156&	 84&	$\bm{799}$\\
   Walk&	 $-42$&	   130&    185& 	126&  $-44$&  $-42$&	 49&	138&	$\bm{500}$\\
  pWalk&	 $-91$&  $-54$&   $-1$&      64&	109&  $-90$&  $-23$&	 76&	$\bm{-10}$\\
    SCT&     $-41$&  $-16$&  $-36$&   $-47$&  $-43$&  $-69$& $-133$&   $-2$&	$\bm{-387}$\\
    RSP&	$-139$&	$-148$&	$-122$&      17&  $-67$& $-166$& $-194$& $-162$&	$\bm{-981}$\\
   Heat&	 $-31$&	$-339$&	$-515$&	 $-513$& $-148$& $-123$& $-458$& $-505$&	$\bm{-2632}$\\
  SP-CT&	$-399$&	$-365$&	$-250$&	 $-186$& $-469$& $-450$& $-414$& $-366$&	$\bm{-2899}$\\
    For&	$-553$&	$-586$&	$-585$&	 $-587$& $-596$& $-600$& $-594$& $-595$&	$\bm{-4696}$\\
			\bottomrule
		\end{tabular}
		\\[8pt]\centerline{(a) optimal parameters}
	\end{minipage}\ \ \ \ \
	\begin{minipage}{.49\textwidth}
		\centering
		\begin{tabular}{lrrrrrrrrr}
			\toprule
			\multicolumn{1}{r}{\textbf{Nodes}}         & 100&    100& 100&  100& 200&  200& 200&  200&{\textbf{Sum}}\\
			\multicolumn{1}{r}{\textbf{Classes}}	   &   2&      2&   4&    4&   2&    2&   4&    4&{\textbf{  of}}\\ %\multicolumn{1}{l}
			\multicolumn{1}{r}{$\bm{p_{\mathbf{out}}}$}& 0.1&	0.15& 0.1& 0.15& 0.1& 0.15& 0.1& 0.15&{\textbf{scores}}\\
			\midrule
logComm\!\!&   471&	   563&	   497&	   472&	   440&	   588&	   591&	   590&	$\bm{ 4212}$\\
   SCCT&	   412&	   448&	   446&	   382&	   470&	   450&	   495&	   498&	$\bm{ 3601}$\\
 logFor&	   171&	   242&	   275&	   166&	    88&	   185&	   296&	   210&	$\bm{ 1633}$\\
   Walk&	  $-4$&	   229&	   291&	   268&	    48&	   145&	   226&	   320&	$\bm{ 1523}$\\
  pWalk&	    19&	    45&	   221&	   232&	   113&	   111&	   188&	   217&	$\bm{ 1146}$\\
     FE&	 $-94$&	    91&	    95&	   240&	 $-47$&	    56&	    18&	   152&	$\bm{  511}$\\
logHeat&	   342&	    91&	 $-22$&	$-243$&	   269&	   156&	    61&	$-376$&	$\bm{  278}$\\
    SCT&	 $-21$&	  $-2$&	$-174$&	$-196$&	  $56$&	  $50$&	 $-46$&	  $54$&	$\bm{-279}$\\
   Comm&	 $-40$&	$-191$&	     2&	 $-58$&	    10&	$-213$&	 $-70$&	$-153$&	$\bm{ -713}$\\
  SP-CT&	$-343$&	$-238$&	$-203$&	$-103$&	$-411$&	$-380$&	$-348$&	$-190$&	$\bm{-2216}$\\
    RSP&	$-473$&	$-335$&	$-328$&	 $-60$&	$-426$&	$-365$&	$-366$&	$-222$&	$\bm{-2575}$\\
   Heat&	  $15$&	$-396$&	$-507$&	$-504$&	 $-64$&	$-198$&	$-445$&	$-500$&	$\bm{-2599}$\\
    For&	$-455$&	$-547$&	$-593$&	$-596$&	$-546$&	$-585$&	$-600$&	$-600$&	$\bm{-4522}$\\
			\bottomrule
		\end{tabular}
		\\[8pt]\centerline{(b) 90th percentiles}
	\end{minipage}
	\caption{\label{t_CopComp}Copeland's scores of the measure families on random graphs}
}\end{table}

%The second competition is more in favor of Walk ($+1$ place)
One can notice a number of differences between the orders of families provided by the first competition and the second one. However, in both cases, logarithmic measures outperform the corresponding plain ones. In particular, it can be observed that FE is also a kind of logarithmic measure, as distinct from~RSP.

Here, the undisputed leader is logComm. Second place goes to SCCT, a measure which is not logarithmic, but involves even a more smoothing sigmoid transformation.

\section{Reject curves}
\label{s_reject}

In this section, we compare the performance of distances (corresponding to the proximity measures or defined independently) in clustering tasks using reject curves.

\subsection{Definition}

The ROC curve (also referred to as the reject curve) for this type of data can be defined as follows.

\begin{itemize}
	\item Let us order the distances $d(x,y),\,$ $x,y\in V(G)$ from the minimum to the maximum, where the distance $d(\cdot,\cdot)$ corresponding to a kernel $p(\cdot,\cdot)$ is produced by the $d(x,y)= p(x,x)+p(y,y)-p(x,y)-p(y,x)$ transformation\footnote{Recall that a number of distances that correspond to logarithmic measures possess a meaningful cutpoint additivity property~\cite{Che13Paris}.}.
	\item To each $d(x,y)$ we assign a point in the $[0,1]\times[0,1]$ square. Its $X$-coordinate is the share of inter-class distances that are less than or equal to $d(x,y),$ the $Y$-coordinate being the share of intra-class distances (between different nodes) that are less than or equal to $d(x,y).$
   \item The polygonal line connecting the consecutive points is the \emph{reject curve\/} corresponding to the graph. \x{^}
%	The reject curve is the dependence of the ``percentage of intra-class distances under the threshold'' upon the ``percentage of inter-class distances under the threshold'' collected for all values of the threshold from the grid.
\end{itemize}

A better measure is characterized by a reject curve that goes higher or, at least, has a larger area under the curve.

\subsection{Results}

%\begin{enumerate}
The optimal values of the family parameters (adjusted to the $[0,\,1]$ segment) w.r.t. the ARI in clustering based on Ward's method for three $G(N, (m)p_\text{in}, p_\text{out})$ models are presented in Table~\ref{t_Optt}.
		\begin{table}[H]
			\begin{tabular}{lrrrr}
				\toprule
				Measure & $G(100, (2)0.3, 0.05)$ & $G(100, (2)0.3, 0.1)$ & $G(100, (2)0.3, 0.15)$\\
                (kernel)& Opt. parameter, ARI    & Opt. parameter, ARI   & Opt. parameter, ARI\\
				\midrule
				pWalk 	& 0.86,\;\;	    0.9653	&0.90,\;\;	    0.8308	&0.66,\;\;	    0.5298\\
				Walk	& 0.86,\;\;	    0.9664	&0.74,\;\;	    0.8442	&0.64,\;\;	    0.5357\\
				For 	& 1.00,\;\;	    0.5816	&0.98,\;\;	    0.3671	&0.00,\;\;	    0.2007\\
				logFor 	& 0.62,\;\;	    0.9704	&0.56,\;\;	    0.8542	&0.52,\;\;	    0.5541\\
				Comm	& 0.38,\;\;	    0.9761	&0.32,\;\;	    0.8708	&0.26,\;\;	    0.5661\\
				logComm & 0.68,\;\;	\ul{0.9838}	&0.54,\;\;	\ul{0.9466}	&0.62,\;\;	\ul{0.7488}\\
				Heat	& 0.86,\;\;	    0.6128	&0.86,\;\;	    0.5646	&0.78,\;\;	    0.2879\\
				logHeat & 0.52,\;\;	    0.9827	&0.40,\;\;	    0.8911	&0.28,\;\;	    0.5561\\
                SCT     & 0.74,\;\;	    0.9651	&0.62,\;\;	    0.8550	&0.64,\;\;	    0.5531\\
                SCCT    & 0.36,\;\;	    0.9834	&0.26,\;\;	    0.9130	&0.22,\;\;	    0.6626\\
				RSP 	& 0.99,\;\;	    0.9712	&0.98,\;\;	    0.8444	&0.98,\;\;	    0.5430\\
				FE 		& 0.94,\;\;	    0.9697	&0.94,\;\;	    0.8482	&0.86,\;\;	    0.5460\\
				SP-CT 	& 0.28,\;\;	    0.9172	&0.34,\;\;	    0.6782	&0.42,\;\;	    0.4103\\
				\bottomrule
			\end{tabular}\caption{\label{t_Optt}Optimal family parameters and the corresponding ARI's}
		\end{table}
		
\Up{.5}
The optimum chosen on the grid of 50 parameter values is shown as the first number in each of three columns. The second number is the ARI corresponding to the optimum averaged over 200 random graphs. The maximum averaged ARI's are underlined. All of them belong to logComm.

\begin{figure}[H] %tb
	\begin{minipage}{.24\textwidth} %49
		\leftfigure{\includegraphics[width=\linewidth]{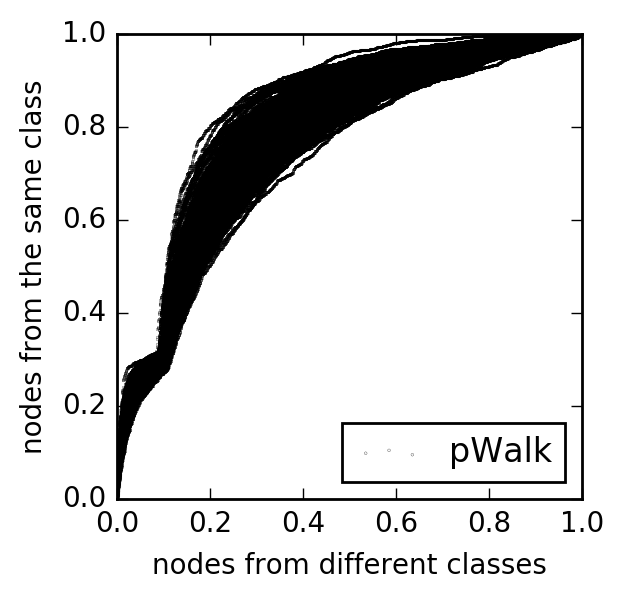}}
		\\\centerline{(a) pWalk}
	\end{minipage}
	\begin{minipage}{.24\textwidth} %49
		\leftfigure{\includegraphics[width=\linewidth]{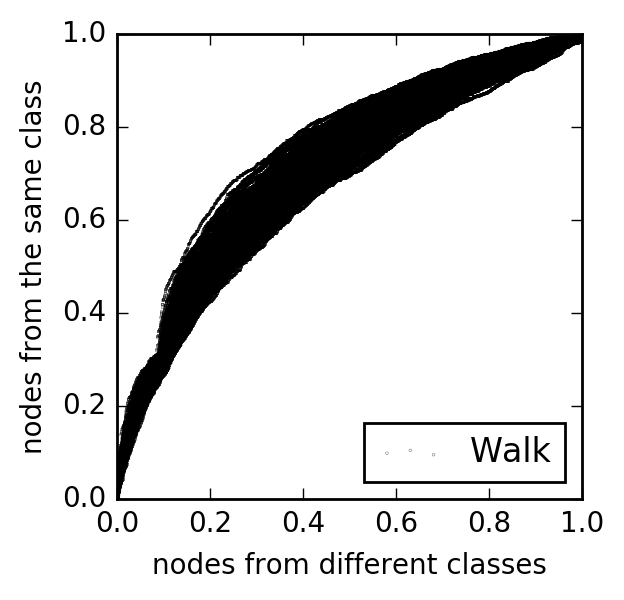}}
		\\\centerline{(b) Walk}
	\end{minipage}
	\begin{minipage}{.24\textwidth} %49
		\leftfigure{\includegraphics[width=\linewidth]{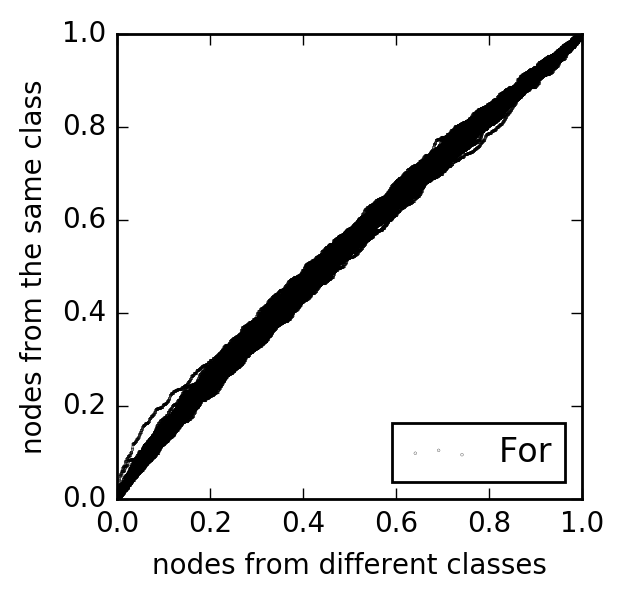}}
		\\\centerline{(c) For}
	\end{minipage}
	\begin{minipage}{.24\textwidth} %49
		\leftfigure{\includegraphics[width=\linewidth]{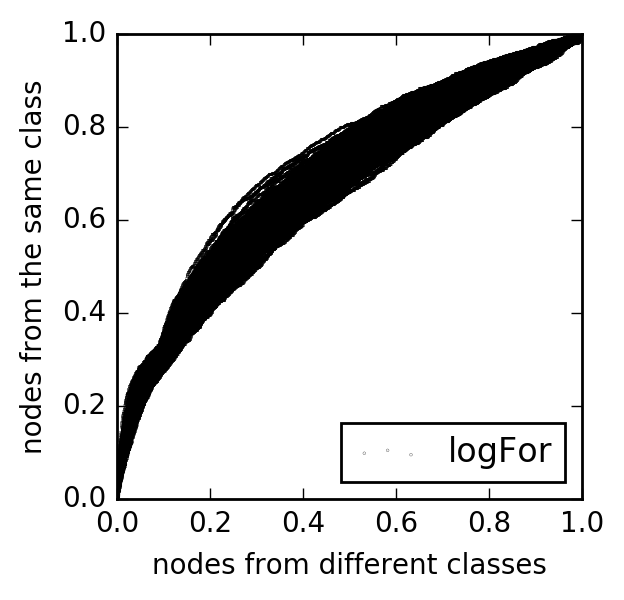}}
		\\\centerline{(d) logFor}
	\end{minipage}
    \\[6pt]
	\begin{minipage}{.24\textwidth} %49
		\leftfigure{\includegraphics[width=\linewidth]{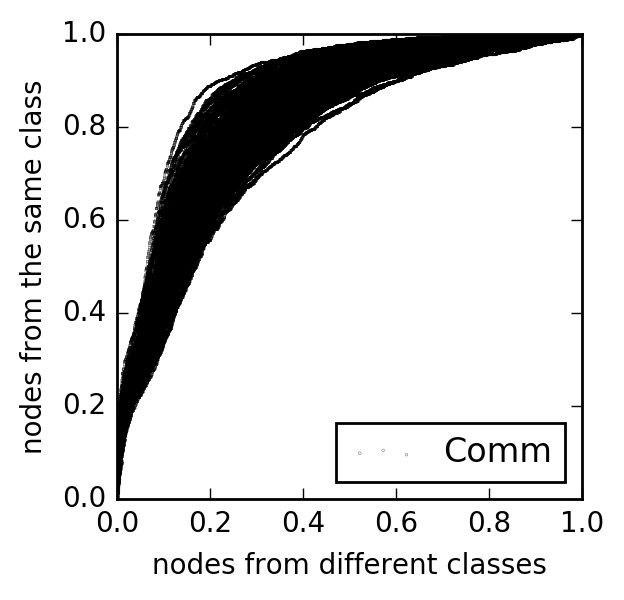}}
		\\\centerline{(e) Comm}
	\end{minipage}
	\begin{minipage}{.24\textwidth} %49
		\leftfigure{\includegraphics[width=\linewidth]{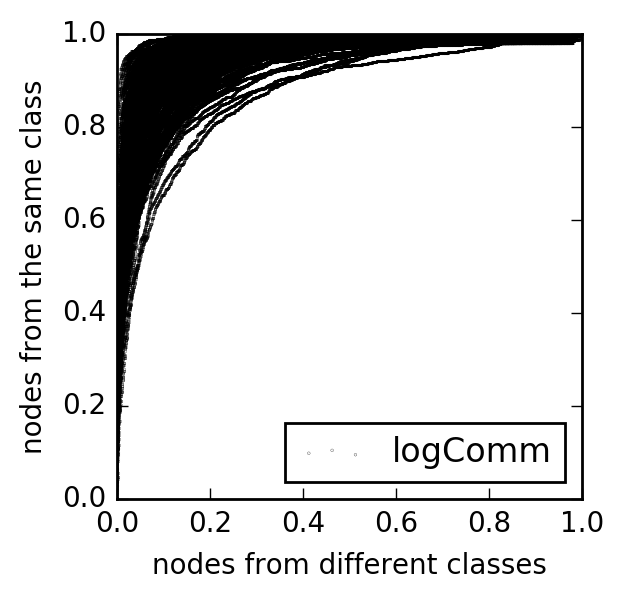}}
		\\\centerline{(f) logComm}
	\end{minipage}
	\begin{minipage}{.24\textwidth} %49
		\leftfigure{\includegraphics[width=\linewidth]{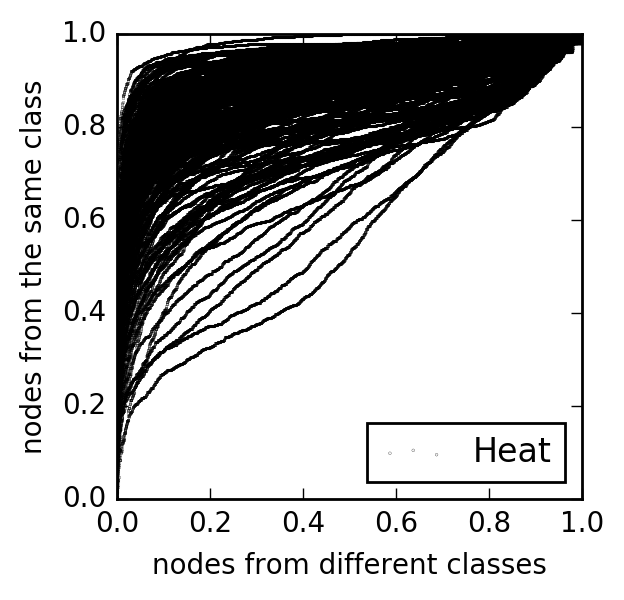}}
		\\\centerline{(g) Heat}
	\end{minipage}
	\begin{minipage}{.24\textwidth} %49
		\leftfigure{\includegraphics[width=\linewidth]{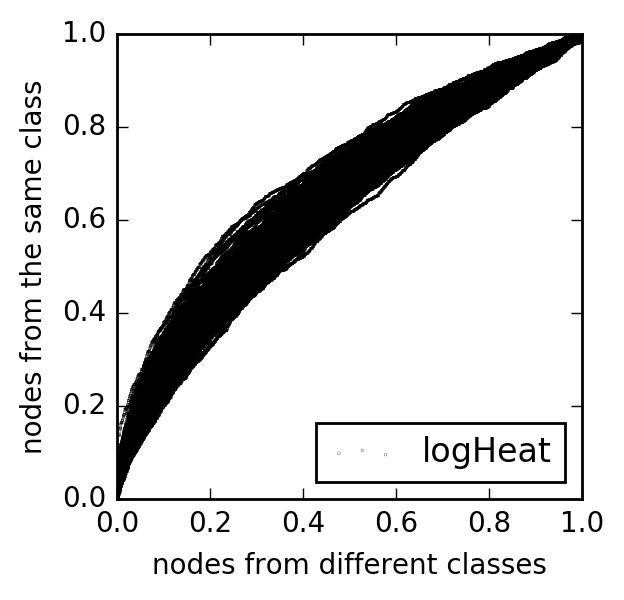}}
		\\\centerline{(h) logHeat}
	\end{minipage}
    \\[6pt]
%   \caption{Reject curves for the graph measures under study (continued)}
%\end{figure}
%\begin{figure}[H] %tb
%    \samenumber
	\begin{minipage}{.196\textwidth} %49
		\leftfigure{\includegraphics[width=\linewidth]{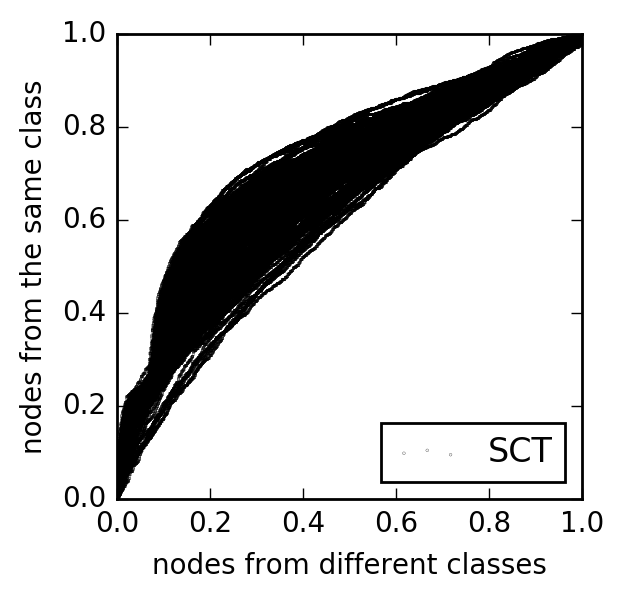}}
		\\\centerline{(i) SCT}
	\end{minipage}
	\begin{minipage}{.196\textwidth} %49
		\leftfigure{\includegraphics[width=\linewidth]{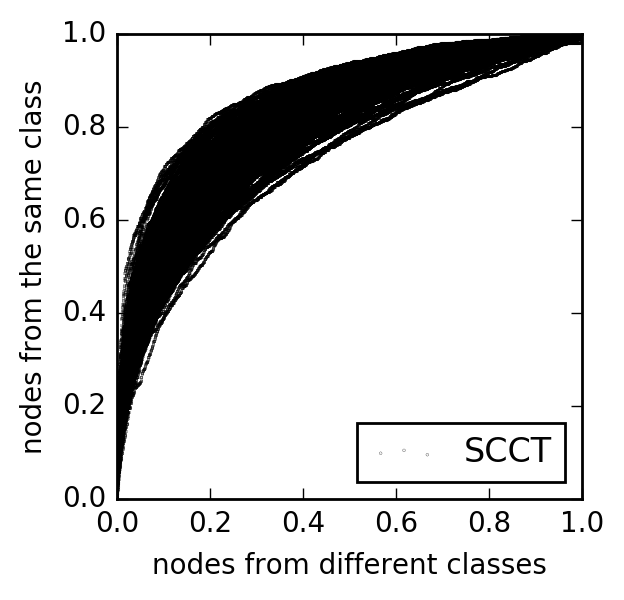}}
		\\\centerline{(j) SCCT}
	\end{minipage}
	\begin{minipage}{.196\textwidth} %49
		\leftfigure{\includegraphics[width=\linewidth]{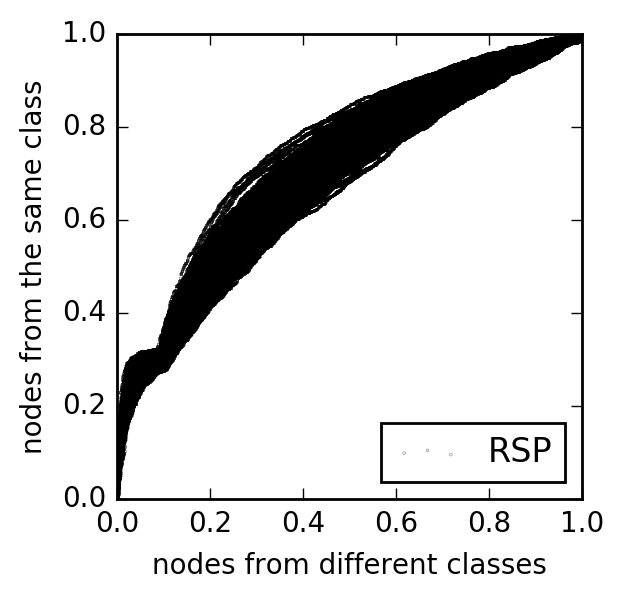}}
		\\\centerline{(k) RSP}
	\end{minipage}
	\begin{minipage}{.196\textwidth} %49
		\leftfigure{\includegraphics[width=\linewidth]{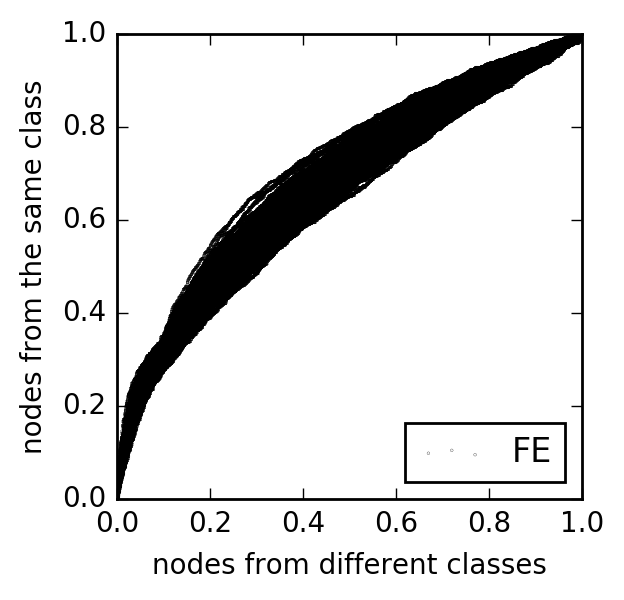}}
		\\\centerline{(l) FE}
	\end{minipage}
	\begin{minipage}{.196\textwidth} %49
		\leftfigure{\includegraphics[width=\linewidth]{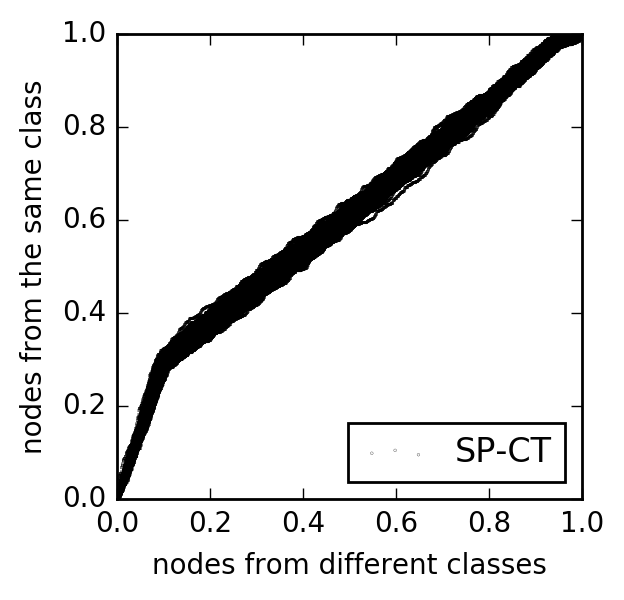}}
		\\\centerline{(m) SP-CT}
	\end{minipage}

    \caption{\label{f_Reject}Reject curves for the graph measures under study}
\end{figure}

The reject curves for $G(100, (2)0.3, 0.1)$ and the optimal values of the family parameters (w.r.t. the ARI of Ward's method clustering) are shown in Fig.\ref{f_Reject}. % The pivot
Each subfigure contains 200 lines corresponding to 200 random graphs.\x{^}

The $\varepsilon$-like bend of several curves (pWalk, Walk, logFor, SCT, RSP, FE) %\x{^}
appears because the corresponding measures strongly correlate with the Shortest path (SP) distance between nodes. In these experiments, the SP distance takes only a few small values.\x{^}
	
\begin{figure}[H] %tb
	\begin{minipage}{.56\textwidth}
		\leftfigure{\includegraphics[width=.75\linewidth]{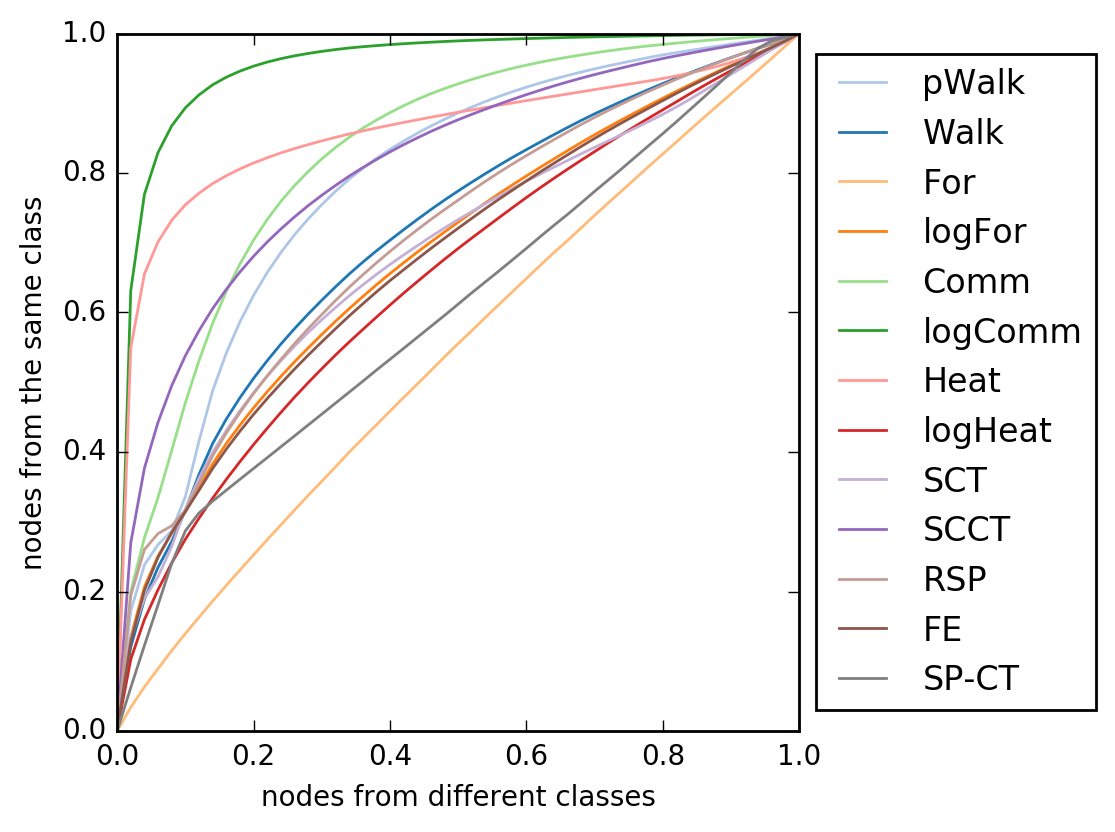}}
		\\\centerline{(a) All families}
	\end{minipage}%
	\begin{minipage}{.42\textwidth}
		\leftfigure{\includegraphics[width=.75\linewidth]{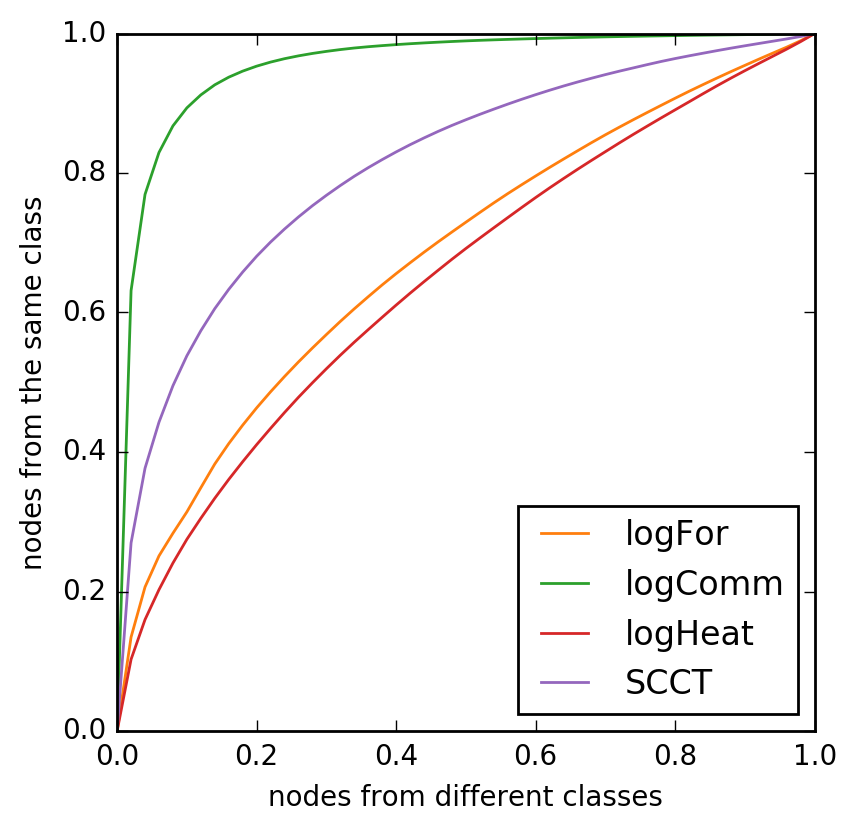}}
		\\\centerline{(b) The four families leading in Table~\ref{t_CopComp}(a)}
	\end{minipage}
\caption{\label{f_Rcur}Average reject curves}
\end{figure}

Finally, in Fig.\,\ref{f_Rcur}(a) we show the reject curves averaged over 200 random graphs. The curves for the four families that are leaders in Table~\ref{t_CopComp} are duplicated in Fig.\,\ref{f_Rcur}(b).
%\end{enumerate}

One can observe that the results are partially concordant with those obtained with Ward's method. In particular, the first place goes to logComm.
\x{^}%, which has a small advantage over SCCT.
Therefore, these results are not an exclusive\x{^} feature of Ward's method. 
Notice that Heat has a good average reject curve. However, it produces relatively many large intra-class distances %poorly distinguishes large inter-class and intra-class distances 
and its partial results are extremely unstable. This supposedly determines the low values of ARI of this measure. \x{^}

\section{Graphs with classes of different sizes}
\label{s_diffsize}

\begin{figure}[H]
	\begin{minipage}{.5\textwidth}
		\leftfigure{\includegraphics[width=.9\linewidth]{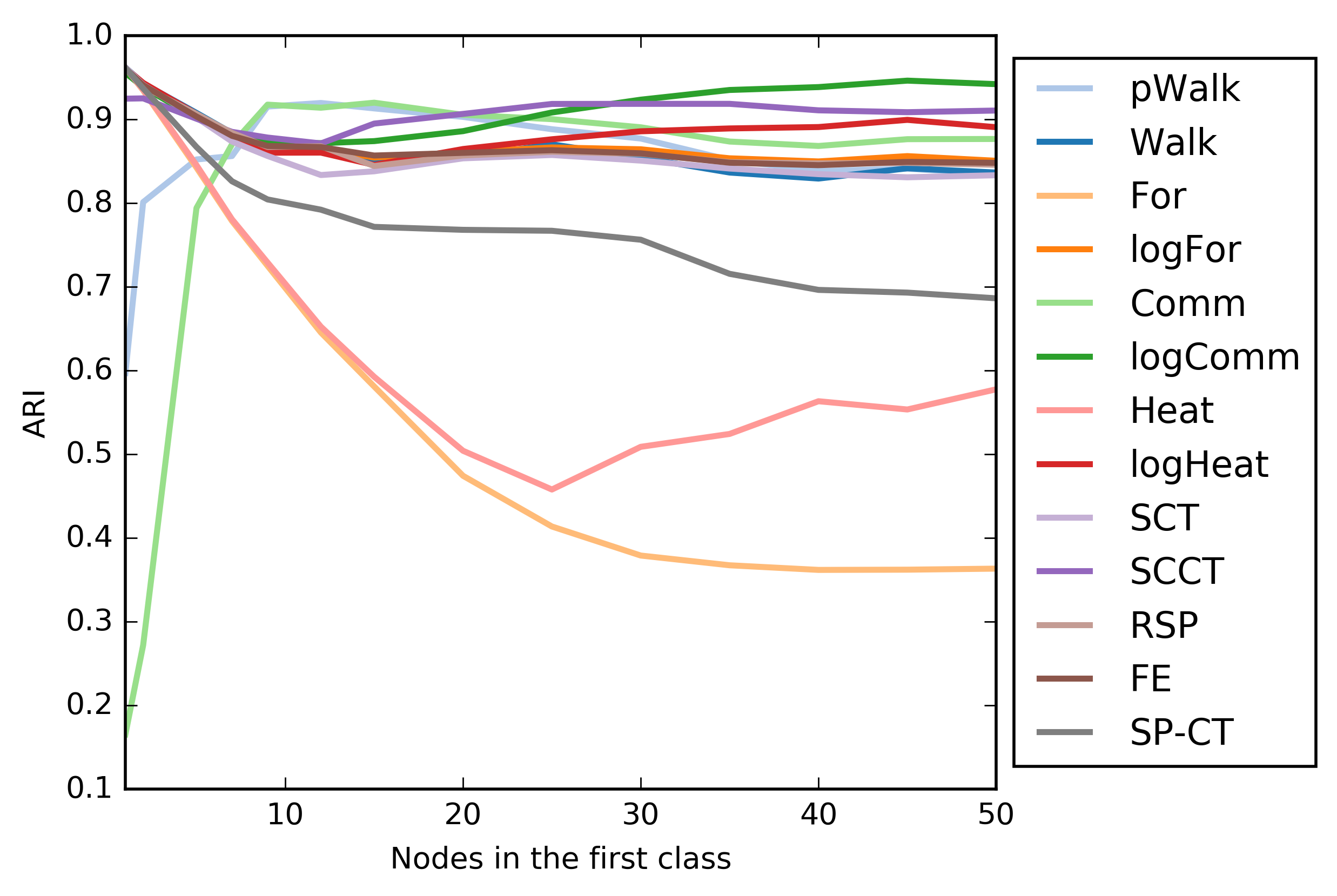}} %0.8
		\\\centerline{(a) All families}
	\end{minipage}%
	\begin{minipage}{.5\textwidth}
		\leftfigure{\includegraphics[width=.9\linewidth]{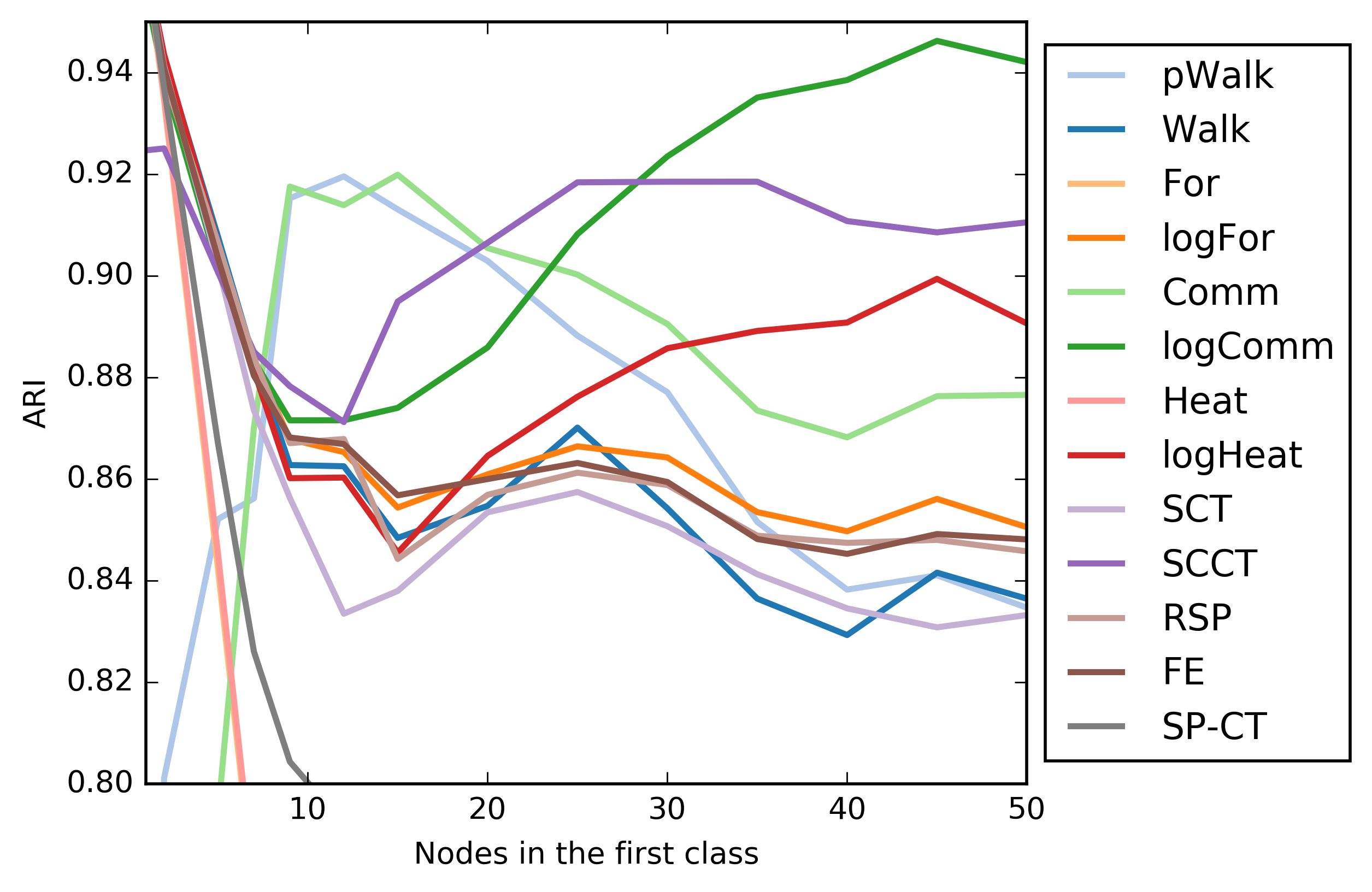}} %1.07
		\\\centerline{(b) Leading families}
	\end{minipage}
\caption{\label{f_difClas}Graphs with two classes of different sizes: clustering with optimal parameter values}
\end{figure}

The $G(N, (m)p_\text{in}, p_\text{out})$ model generates graphs with nodes divided into classes of the same size.
We now consider graphs with $N=100$ nodes divided into two classes of different sizes. The size of the first class, $N_1$, is shown along the horizontal axis in Fig.\,\ref{f_difClas}.

We see that the ARI's of logComm, SCCT, and logHeat have minima at $N_1$ near $10$ or~$15.$ In contrast, the ARI's of Comm and pWalk have larger maxima in the same interval. %at $N_1\approx15.$ \x{^}
As a result, the latter two measures outperform the former three (and the other measures under study) at $N_1\in[8,19].$\x{^} However, if $N_1$ is very small, then Ward's method with Comm or pWalk seems to engender misrecognition. Thus, this case\x{^} can be considered as an exception to the rule that ``logarithmic measures outperform plain ones'': with a moderate size of the smaller class, Comm and pWalk outperform the logarithmic measures\x{^} (and also SCCT in which the sigmoid function is analogous to the logarithmic one as a smoothing transformation\x{^}).

\begin{figure}[H]
%	\leftfigure{\includegraphics[width=.85\linewidth]{{AddFigures/2clusters_fromspark2-ave}.png}}
	\leftfigure{\includegraphics[width=.45\linewidth]{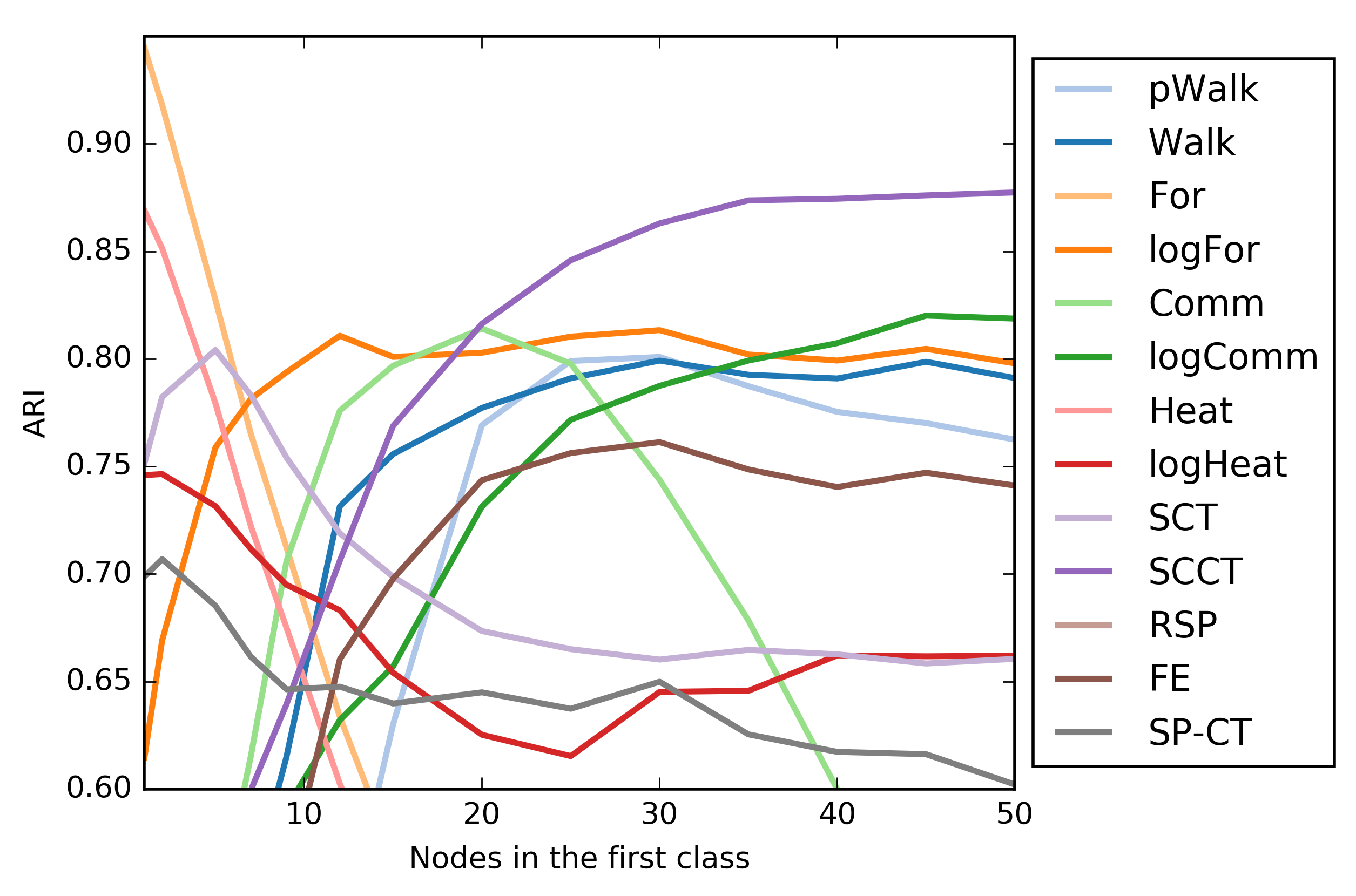}}
%       \\\centerline{}
%\caption{\label{...}...}
\caption{\label{f_difClas1}Graphs with two classes of different sizes: random parameter values}
\end{figure}

In all the above experiments, we looked for the optimal values of the family parameters. If the families of measures are used with random parameter values, then the rating of the families differs. Now,\x{^} %for the classes of the same size
the leader and the vice-leader are SCCT and logFor, respectively, which are most robust to the variation of the family parameter; when one class is very small, the winners are For, SCT, and Heat, see Fig.\,\ref{f_difClas1}.

\Up{2}

%\Up{2}
%\begin{figure}[H]
%\samenumber
%\begin{minipage}{.45\textwidth}
%{\normalsize
%In all the above experiments, we looked for the optimal values of the family parameters. If the families of measures are used with random parameter values, then the rating of the families differs. In the latter case, the leader and the vice-leader are SCCT and logFor, respectively, which are most robust to the variation of the family parameter; when one class is very small, the winners are For, SCT, and Heat, see Fig.\,\ref{f_difClas1}.
%}
%\end{minipage}
%\begin{minipage}{.45\textwidth}
%%	\leftfigure{\includegraphics[width=.85\linewidth]{{AddFigures/2clusters_fromspark2-ave}.png}}
%	\leftfigure{\includegraphics[width=.85\linewidth]{{AddFigures/2clusters_random2}.png}}
%%       \\\centerline{}
%%\caption{\label{...}...}
%\end{minipage}
%\twocaptionwidth{.45\textwidth}{.50\textwidth}\phantom{\rightcaption{}}\rightcaption{\label{f_difClas1}Graphs with two classes of different sizes: random parameter values}
%\end{figure}

\Up{1.5}
Now let us consider a highly heterogeneous data structure: 150 nodes are divided into six classes whose sizes are 65, 35, 25, 13, 8, and~4. The classes are numbered in the descending order of size. The probability of an edge connecting two vertices that belong to classes $i$ and $j$ is the entry $p_{ij}$ of the matrix~$P:$

\Up{2}
\begin{figure}[H]
\samenumber
\begin{minipage}{.45\textwidth}
{\normalsize
$$
P=\begin{pmatrix}
    0.30& 0.20& 0.10& 0.15& 0.07& 0.25\\
    0.20& 0.24& 0.08& 0.13& 0.05& 0.17\\
    0.10& 0.08& 0.16& 0.09& 0.04& 0.12\\
    0.15& 0.13& 0.09& 0.20& 0.02& 0.14\\
    0.07& 0.05& 0.04& 0.02& 0.12& 0.04\\
    0.25& 0.17& 0.12& 0.14& 0.04& 0.40\\
  \end{pmatrix}.
$$}
\end{minipage}
\begin{minipage}{.45\textwidth}
	\leftfigure{\includegraphics[width=.85\linewidth]{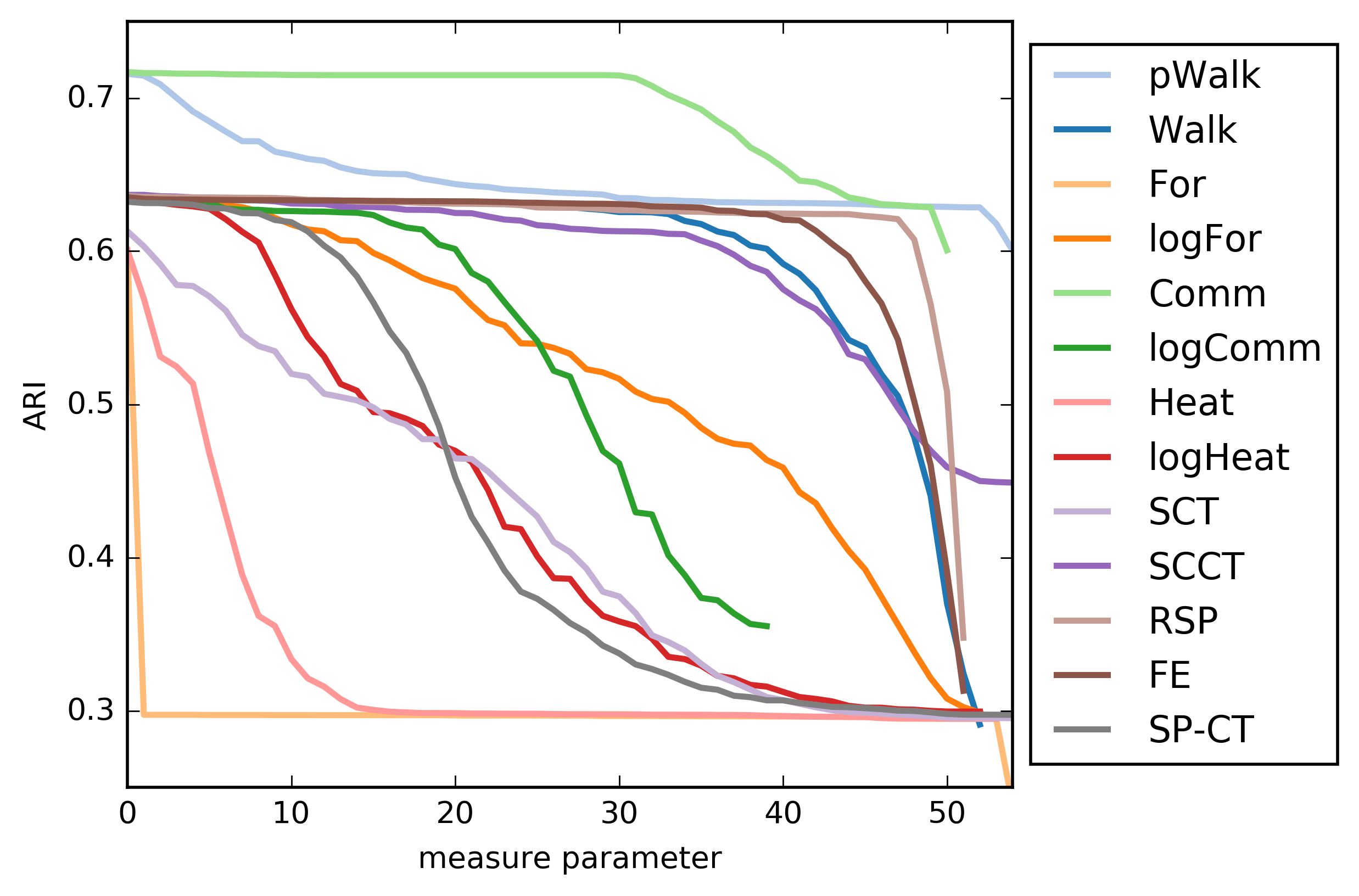}}
%       \\\centerline{}
%\caption{\label{f_6classes}ARI of various measure families on a data structure with 6 classes}
\end{minipage}
\twocaptionwidth{.45\textwidth}{.45\textwidth}\phantom{\rightcaption{}}\rightcaption{\label{f_6classes}ARI of various measure families on a structure with 6 classes}
\end{figure}

\Up{1.5}
For each measure family, we considered 55 values of the family parameter and sorted them in the descending order of the corresponding ARI averaged for 200 random graphs. ARI against the rank of the family parameter value is shown in Fig.\,\ref{f_6classes}. Two things are important for each family: first, the maximum of ARI and second, the velocity of descent.

For this data structure, the leaders are Comm and pWalk, as well as for the two-component graphs with one small, but not very small class of nodes.
%Among the logarithmic measures, the best one is logHeat.

\section{Cluster analysis on several classical datasets}
\label{s_classic}

Hitherto we mainly considered one type of random graph: the graphs with uniform interclass edge probabilities and uniform intraclass edge probabilities. Certainly, many real-world graphs can hardly be obtained in the framework of that model. In this section, we study clustering on several datasets frequently used to check various graph algorithms.

We investigate a total of 9 graphs, the smallest of which (Zachary's Karate club \cite{zachary1977information}) contains 34 nodes. The largest graph (a Newsgroup graph \cite{YenFoussSaerens09} with three classes) contains 600 nodes. We analyse six Newsgroup datasets. The remaining datasets are Football \cite{girvan2002community} and Political books \cite{newman2006modularity}. Table \ref{table:datasets} presents some features of the datasets.
\begin{table}[H]
	\centering
	\begin{tabular}{|c|c|c|c|}
		\hline
		Dataset family & Dataset name & Number of nodes & Number of classes \\
		\hline
		Football & football & 115 & 12 \\
		\hline
		Political books & polbooks & 105 & 3 \\
		\hline
		Zachary & Zachary & 34 & 2 \\
		\hline
		\multirow{6}{*}{Newsgroup} & news\_2cl\_1 & 400 & 2 \\
		\cline{2-4}
		 & news\_2cl\_2 & 400 & 2 \\
		\cline{2-4}
		 & news\_2cl\_3 & 400 & 2 \\
		\cline{2-4}
		 & news\_3cl\_1 & 600 & 3 \\
		\cline{2-4}
		 & news\_3cl\_2 & 600 & 3 \\
		\cline{2-4}
		 & news\_3cl\_3 & 600 & 3 \\
		\hline
	\end{tabular}
	\caption{Overview of the datasets in the experiments\label{table:datasets}}
\end{table}

\begin{figure}[H]
	\begin{minipage}{.33\textwidth}
		\leftfigure{\includegraphics[width=\linewidth]{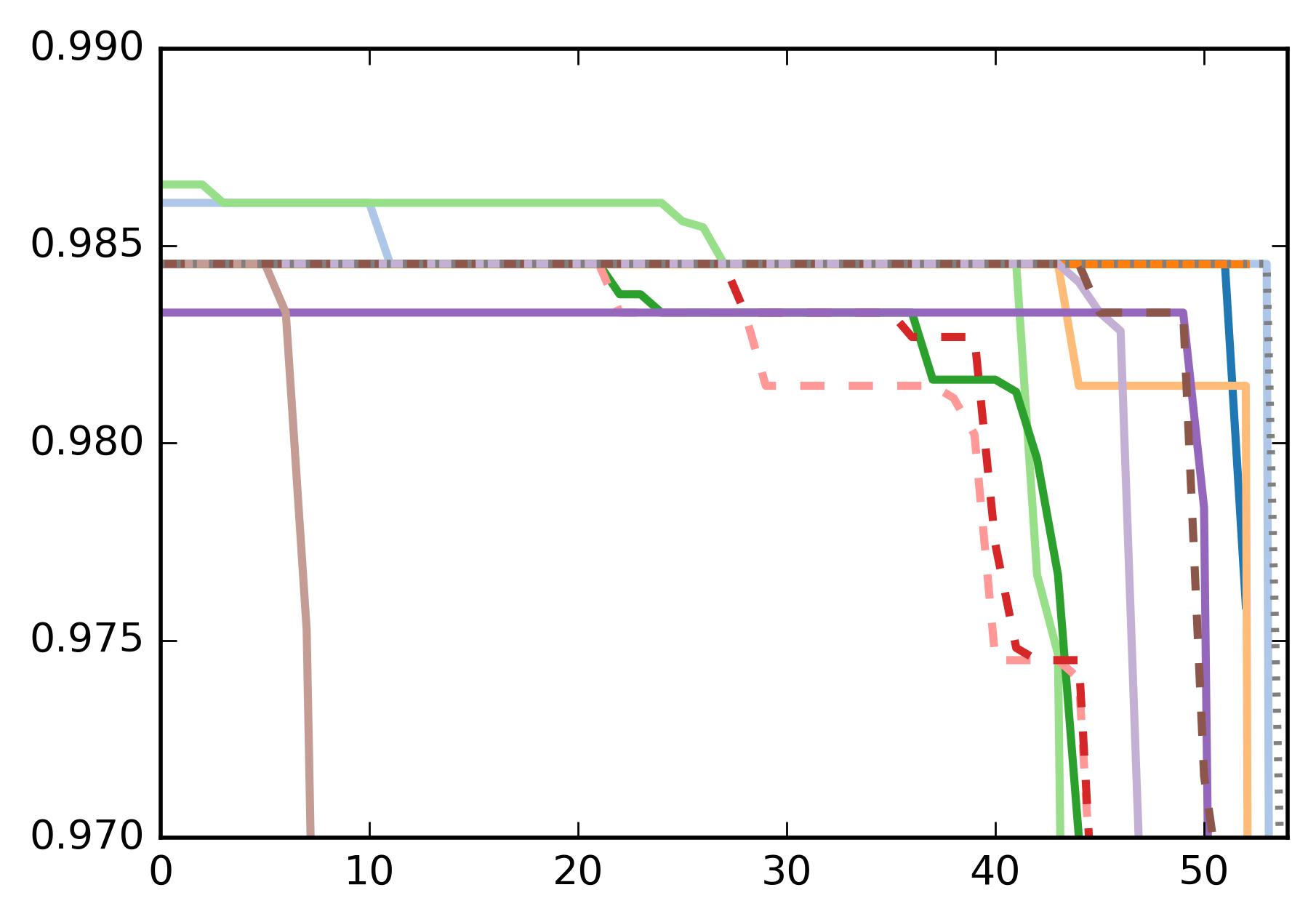}}
		\\\centerline{(a) football}
	\end{minipage}
	\begin{minipage}{.33\textwidth}
		\leftfigure{\includegraphics[width=\linewidth]{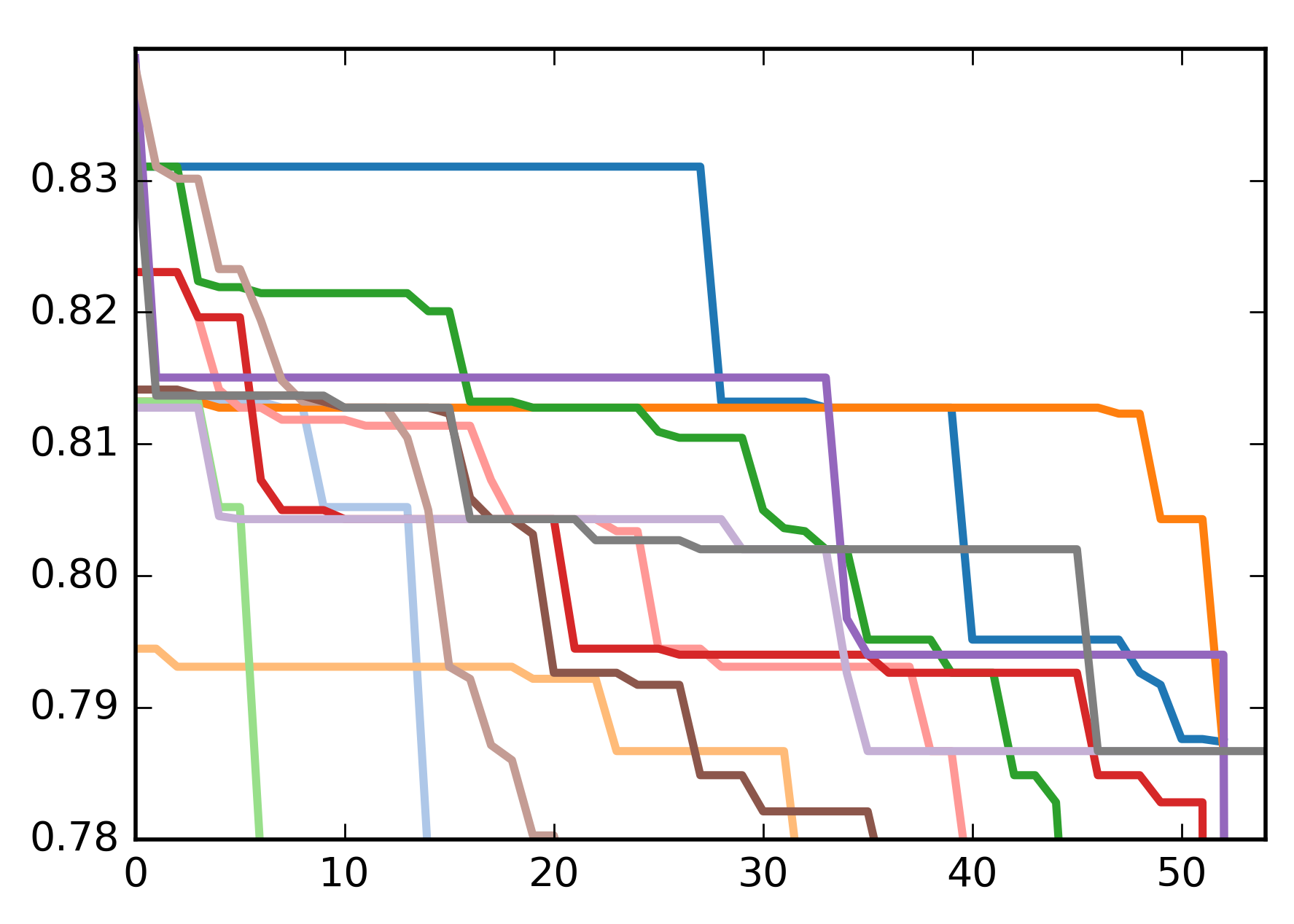}}
		\\\centerline{(b) polbooks}
	\end{minipage}
	\begin{minipage}{.33\textwidth}
		\leftfigure{\includegraphics[width=\linewidth]{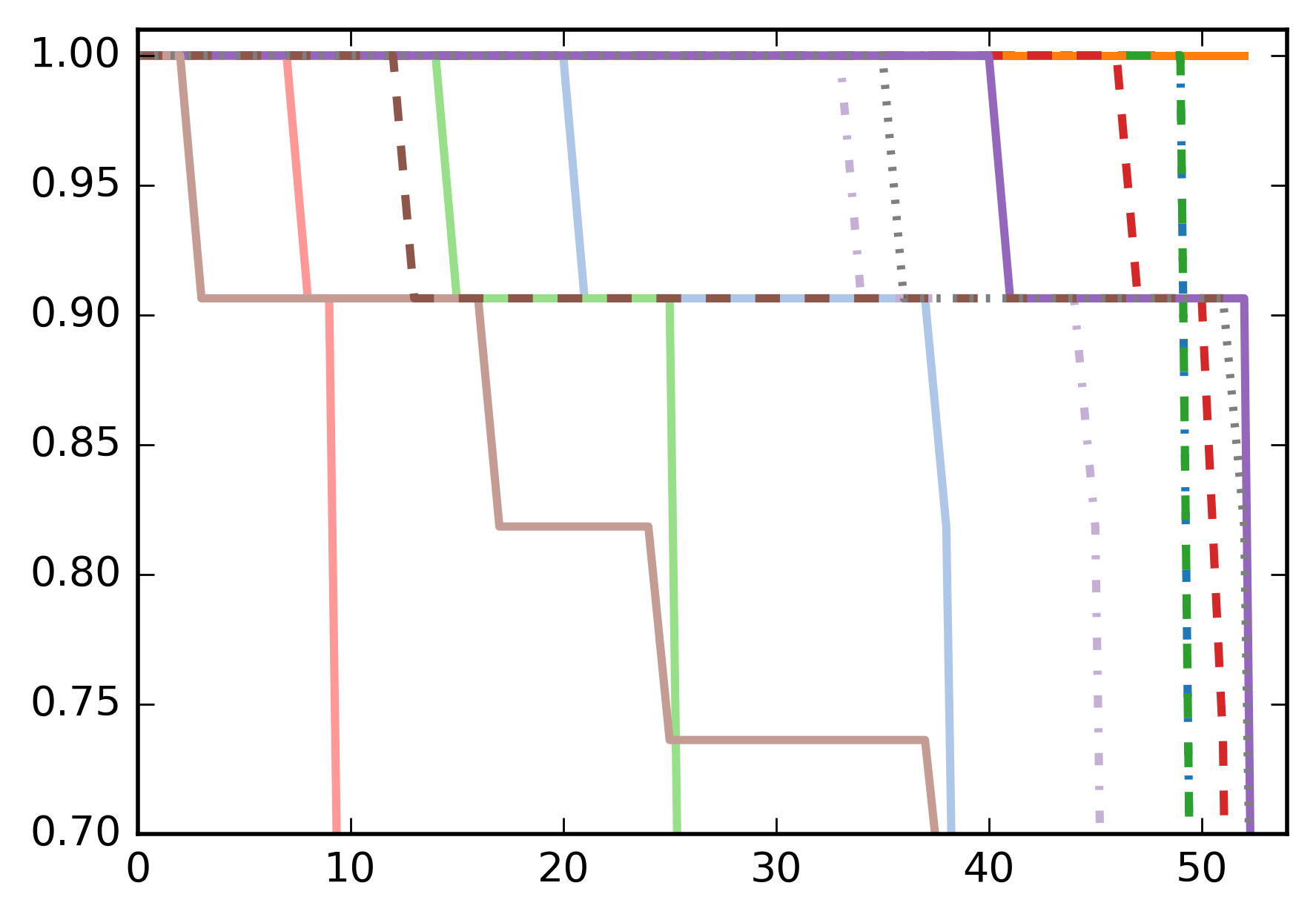}}
		\\\centerline{(c) Zachary}
	\end{minipage}
    \\[10pt]
	\begin{minipage}{.33\textwidth}
		\leftfigure{\includegraphics[width=\linewidth]{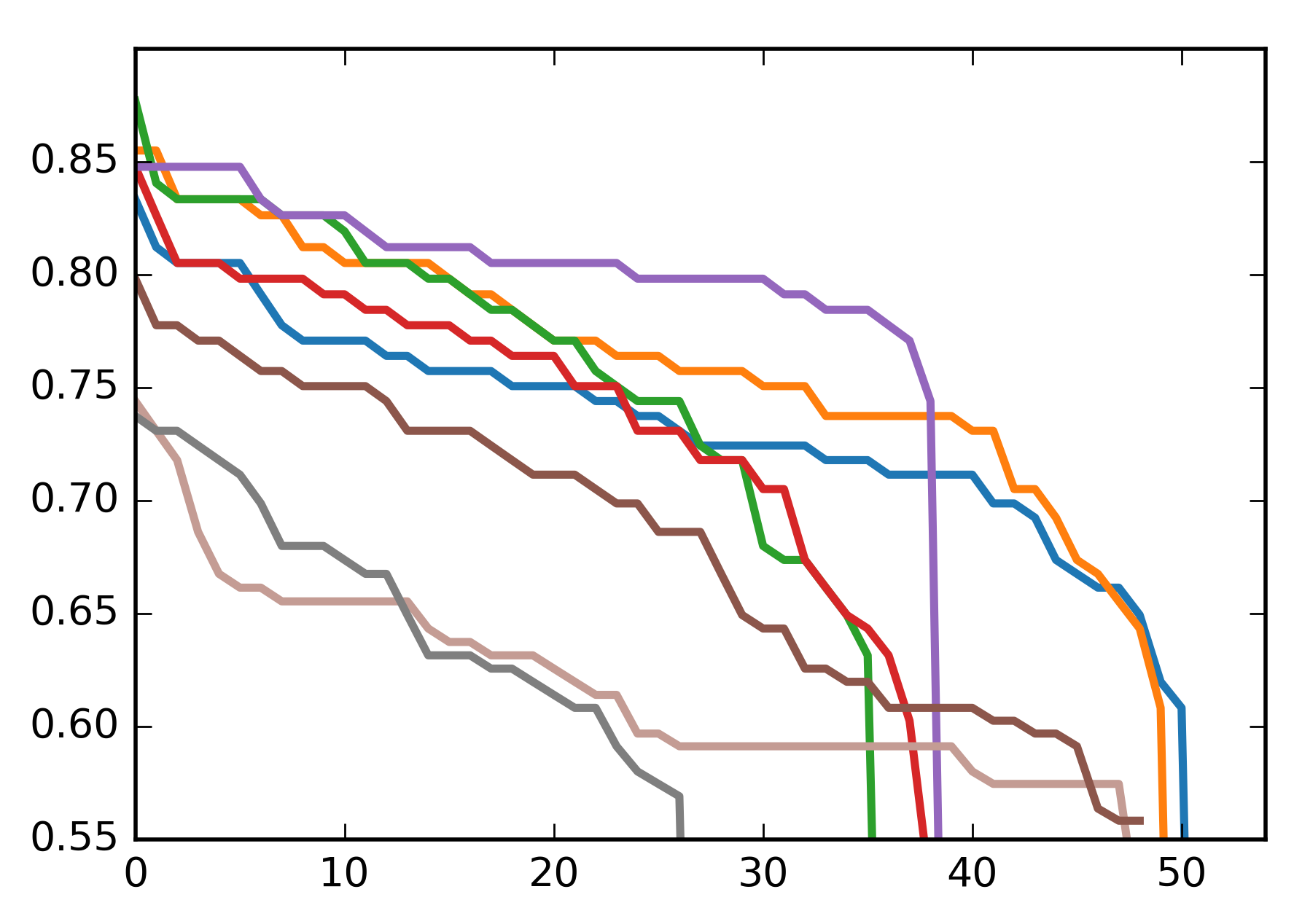}}
		\\\centerline{(d) news\_2cl\_1}
	\end{minipage}
	\begin{minipage}{.33\textwidth}
		\leftfigure{\includegraphics[width=\linewidth]{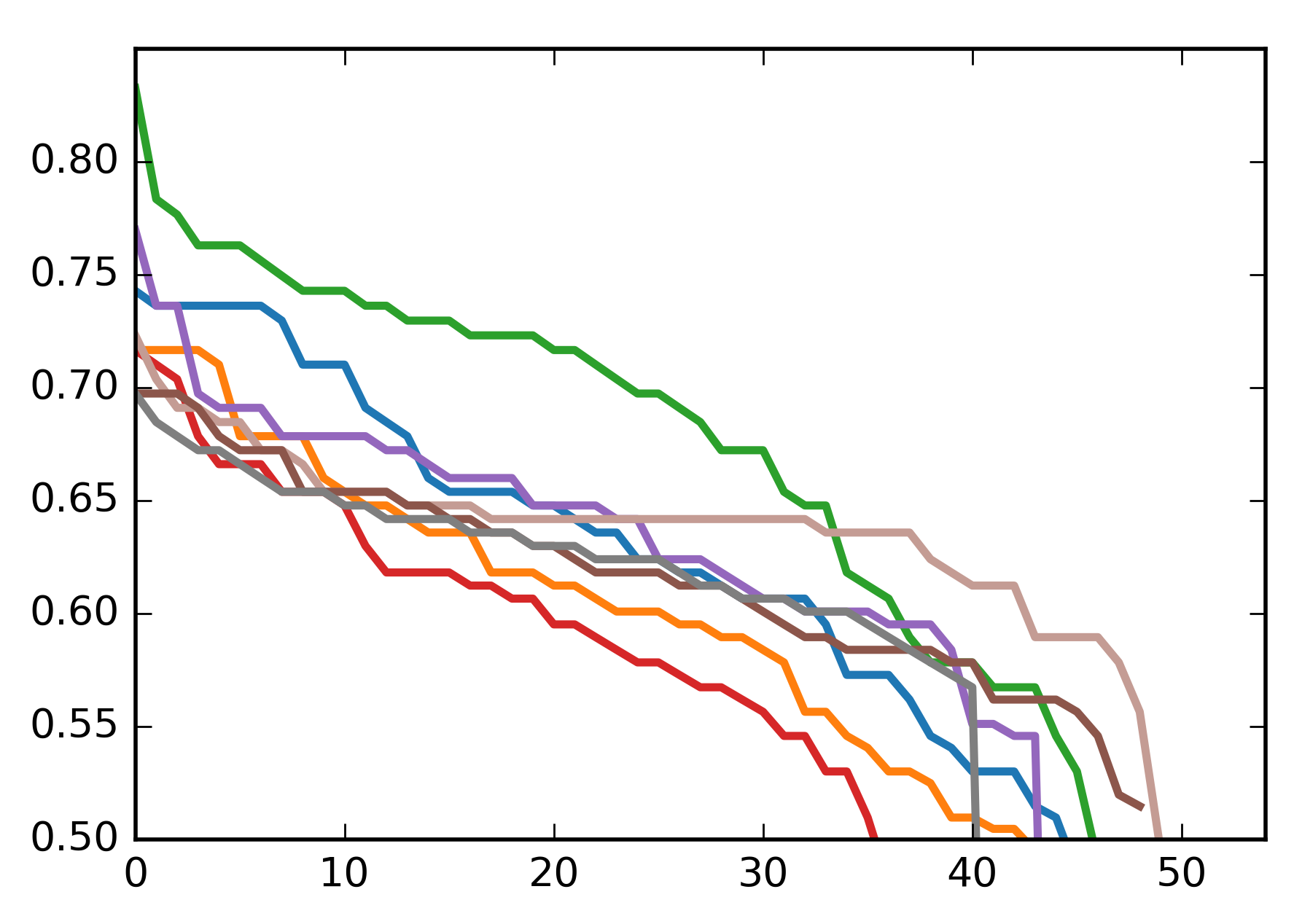}}
		\\\centerline{(e) news\_2cl\_2}
	\end{minipage}
	\begin{minipage}{.33\textwidth}
		\leftfigure{\includegraphics[width=\linewidth]{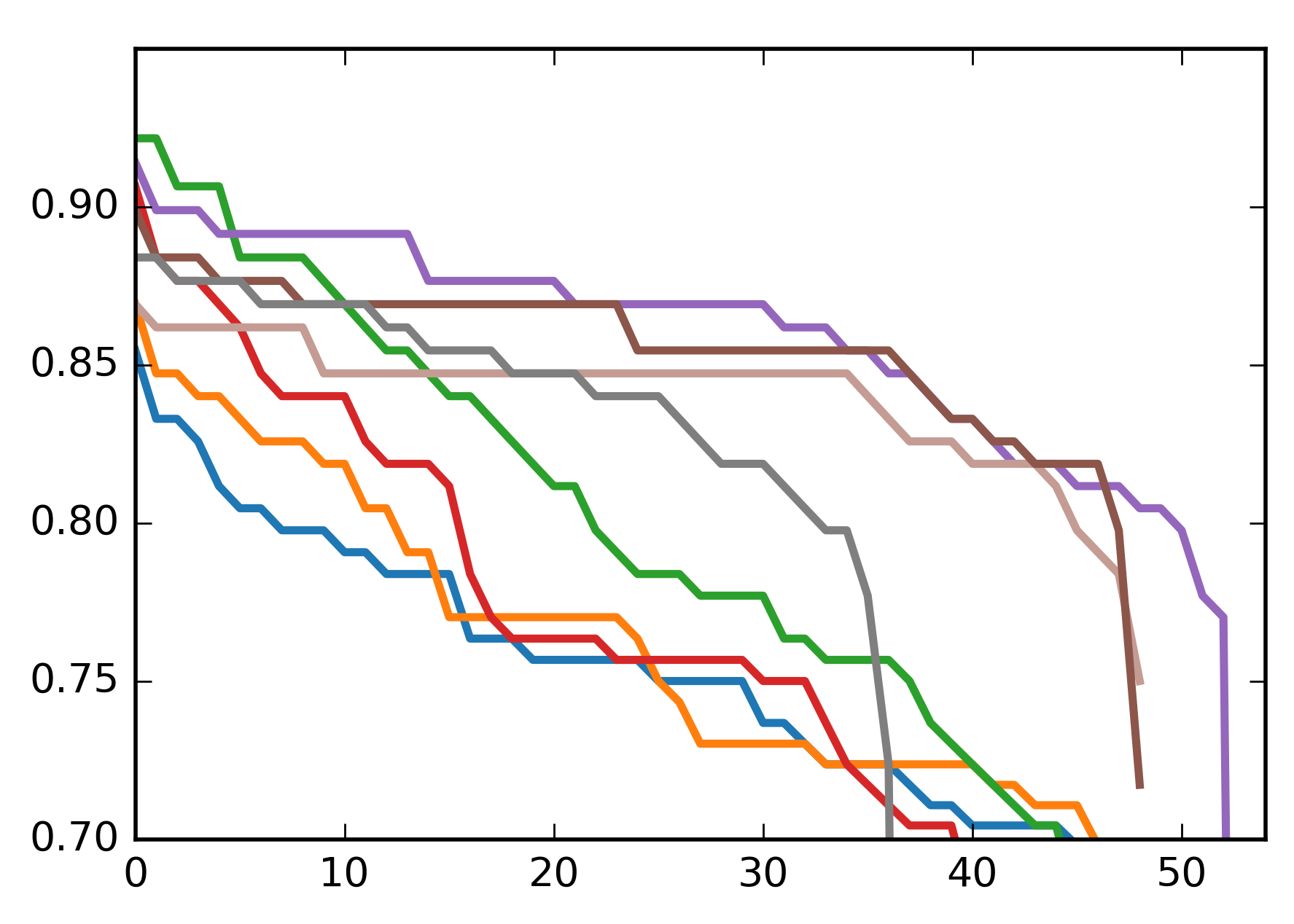}}
		\\\centerline{(f) news\_2cl\_3}
	\end{minipage}
    \\[10pt]
	\begin{minipage}{.33\textwidth}
		\leftfigure{\includegraphics[width=\linewidth]{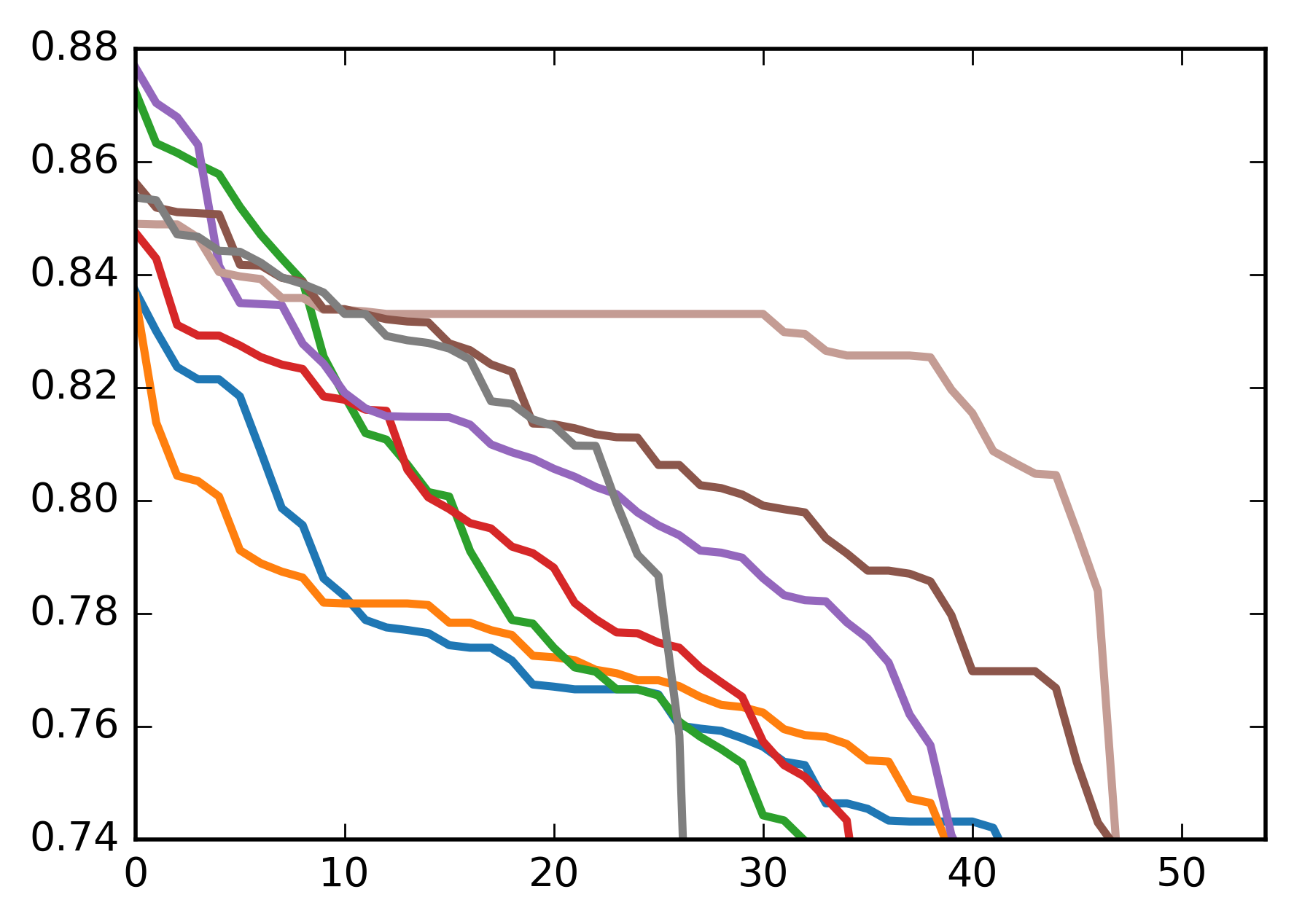}}
		\\\centerline{(g) news\_3cl\_1}
	\end{minipage}
	\begin{minipage}{.33\textwidth}
		\leftfigure{\includegraphics[width=\linewidth]{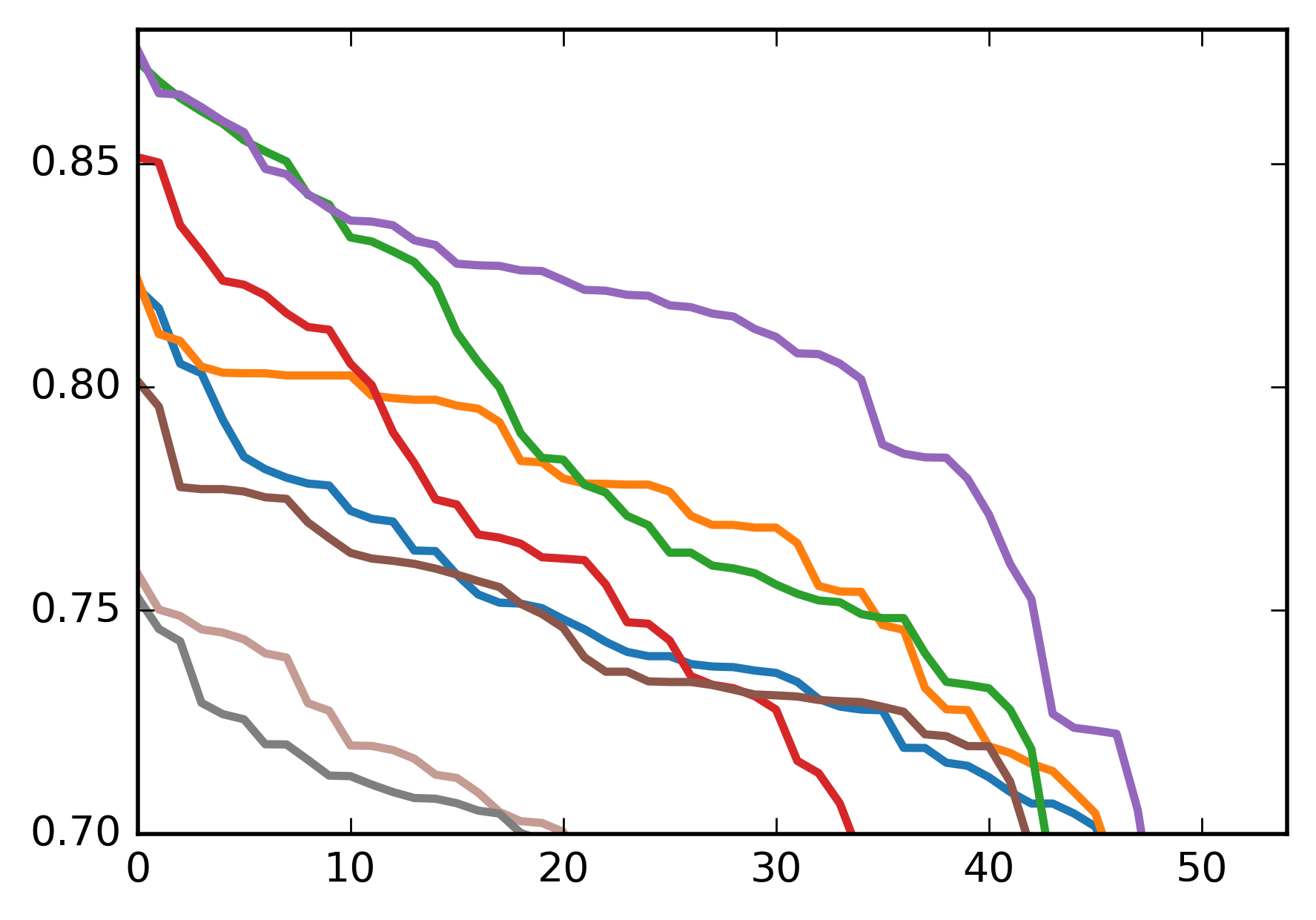}}
		\\\centerline{(h) news\_3cl\_2}
	\end{minipage}
	\begin{minipage}{.33\textwidth}
		\leftfigure{\includegraphics[width=\linewidth]{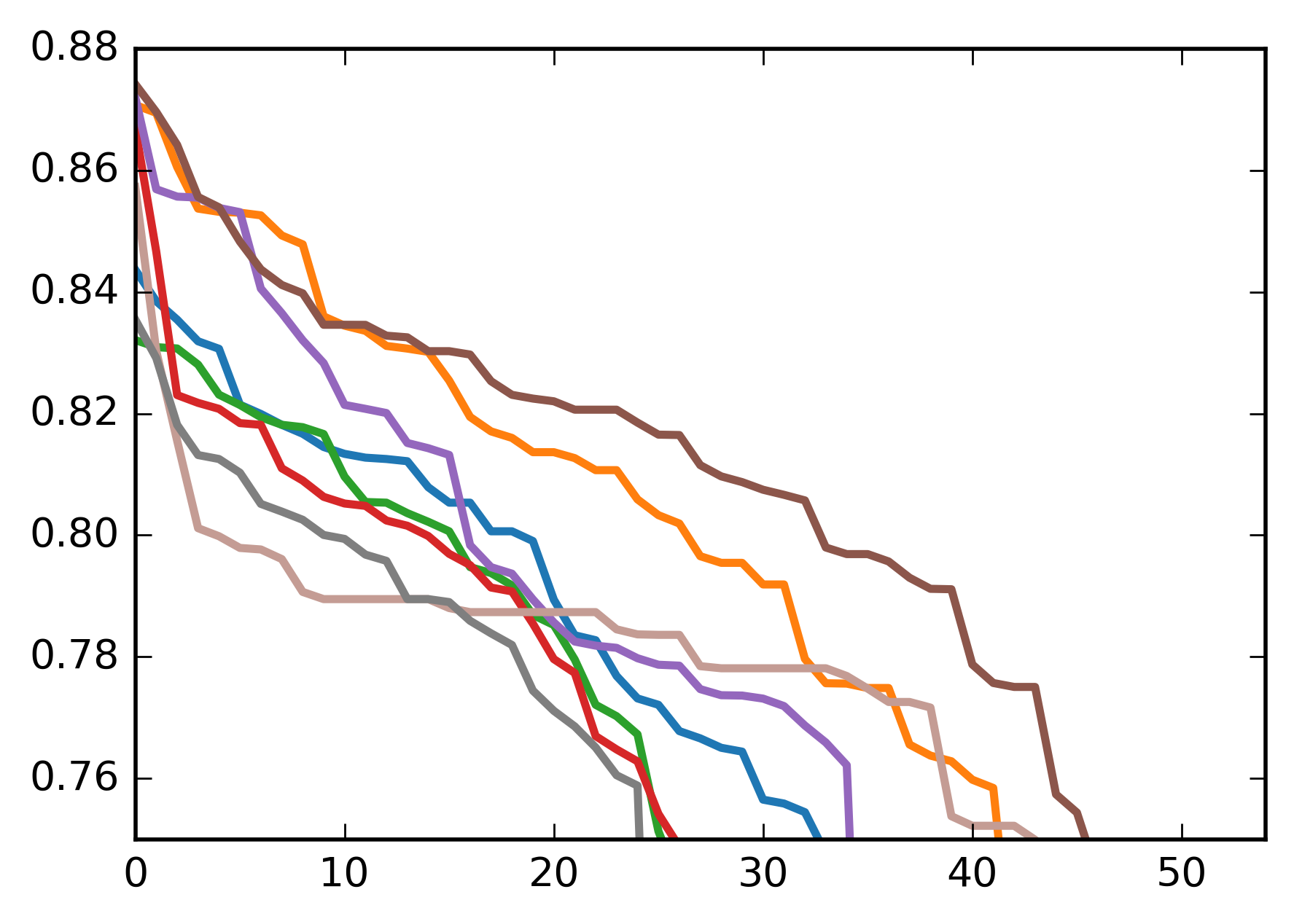}}
		\\\centerline{(i) news\_3cl\_3}
	\end{minipage}
	\\[10pt]
    \begin{minipage}{\textwidth}
        \hfill\includegraphics[width=0.7\linewidth]{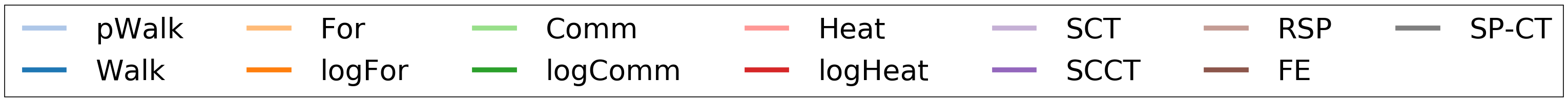}\hfill
%		\rightfigure{\includegraphics[width=0.7\linewidth]{AddFigures/legend.png}}
	\end{minipage}
  \caption{\label{f_datasets}ARI of various measure families on classical datasets}
\end{figure}

\Up{1}
%For each dataset and each measure family, we considered 55 values of the family parameter and sorted them in the descending order of the corresponding ARI. ARI against the rank of the family parameter value is shown in Fig.\,\ref{f_datasets}. Two things are important for each family: first, the maximum of ARI and second, the velocity of descent.
For each dataset and each measure family, we sorted 55 values of the family parameter in the descending order of the corresponding ARI. ARI against the rank of the family parameter value is shown in Fig.\,\ref{f_datasets}.

\Up{2}

\Up{2}
Finally, we present Copeland's score competition for the measure families: separately for the best values of the family parameters and for 80th percentiles (Tables~\ref{t_CopComp2} and~\ref{t_CopComp3}).

\Up{2}
\begin{table}[H] %[H]
	\begin{minipage}{.99\textwidth}		
		\centering
		$$\begin{array}{lrrrrrrrrrr}
			\toprule		
            &\text{ football}&\text{ polbooks}&\text{ Zachary}&\text{ news\_2cl\_1}&\text{ news\_2cl\_2}&
            \text{ news\_2cl\_3}&\text{ news\_3cl\_1}&\text{ news\_3cl\_2}&\text{ news\_3cl\_3}&\textbf{ Score}\\
			\midrule
            \mbox{SCCT}    & -12&  12 &   1 &   7 &  10 &  10 &  12 &  12 &  10 & \bm{62}\\
            \mbox{logComm} & -1 &   5 &   1 &  12 &  12 &  12 &  10 &  10 &  -2 & \bm{59}\\
            \mbox{logHeat} & -1 &   1 &   1 &   7 &   3 &   8 &   2 &   8 &   6 & \bm{35}\\
            \mbox{FE}      & -1 &  -2 &   1 &   2 &  -1 &   6 &   8 &   0 &  12 & \bm{25}\\
            \mbox{RSP}     & -1 &  10 &   1 &   0 &   6 &   1 &   4 &  -2 &   4 & \bm{23}\\
            \mbox{Walk}    & -1 &   5 &   1 &   4 &   8 &  -4 &   0 &   4 &   2 & \bm{19}\\
            \mbox{logFor}  & -1 &  -6 &   1 &  10 &   3 &   1 &  -4 &   6 &   8 & \bm{18}\\
            \mbox{SP-CT}   & -1 &   8 &   1 &  -3 &  -1 &   4 &   6 &  -4 &   0 & \bm{10}\\
            \mbox{SCT}     & -1 & -10 &   1 &  -3 &  -4 &  -2 &  -2 &   2 &  -4 & \bm{-23}\\
            \mbox{Comm}    & 12 &  -6 &   1 &  -6 &  -6 &  -6 &  -6 &  -8 &  -8 & \bm{-33}\\
            \mbox{pWalk}   & 10 &  -6 &   1 &  -8 &  -8 &  -8 &  -8 &  -6 &  -6 & \bm{-39}\\
            \mbox{Heat}    & -1 &   1 &   1 & -10 & -10 & -11 & -11 & -10 & -11 & \bm{-62}\\
            \mbox{For}     & -1 & -12 & -12 & -12 & -12 & -11 & -11 & -12 & -11 & \bm{-94}\\
			\bottomrule
		\end{array}$$
		%\\\centerline{(a) optimal parameters}
	\end{minipage}
	\caption{\label{t_CopComp2}Copeland's scores of the measure families for the best parameter values}
\end{table}

\begin{table}[H] %[H]
	\begin{minipage}{.99\textwidth}
		\centering
		$$\begin{array}{lrrrrrrrrrr}
			\toprule
            &\text{ football}&\text{ polbooks}&\text{ Zachary}&\text{ news\_2cl\_1}&\text{ news\_2cl\_2}&
            \text{ news\_2cl\_3}&\text{ news\_3cl\_1}&\text{ news\_3cl\_2}&\text{ news\_3cl\_3}&\textbf{ Score}\\
			\midrule
        \mbox{logComm} &   0 &  10 &   3 &  10 & 12 &  8 &   4 &  10 &   4 & \bm{61}\\
        \mbox{SCCT}    & -10 &   8 &   3 &  12 &  8 & 12 &   6 &  12 &   8 & \bm{59}\\
        \mbox{FE}      &   0 &   3 &   3 &   2 &  4 &  8 &  12 &   2 &  12 & \bm{46}\\
        \mbox{Walk}    &   0 &  12 &   3 &   4 & 10 & -4 &   0 &   4 &   6 & \bm{35}\\
        \mbox{logFor}  &   0 &   3 &   3 &   8 &  4 & -2 &  -2 &   6 &  10 & \bm{30}\\
        \mbox{SP-CT}   &   0 &   3 &   3 &   0 & -1 &  8 &   8 &  -2 &   0 & \bm{19}\\
        \mbox{logHeat} &   0 &  -7 &   3 &   6 & -1 &  2 &   2 &   8 &   2 & \bm{15}\\
        \mbox{RSP}     & -12 &   3 &  -8 &  -2 &  4 &  4 &  10 &   0 &  -2 & \bm{-3}\\
        \mbox{SCT}     &   0 &  -7 &   3 &  -8 & -4 &  0 &  -4 &  -4 &  -4 & \bm{-28}\\
        \mbox{pWalk}   &  11 &  -4 &   3 &  -6 & -8 & -8 &  -6 &  -7 &  -6 & \bm{-31}\\
        \mbox{Comm}    &  11 & -12 &   3 &  -4 & -6 & -6 &  -8 &  -7 &  -8 & \bm{-37}\\
        \mbox{Heat}    &   0 &  -2 & -11 & -11 &-11 &-11 & -11 & -11 & -11 & \bm{-79}\\
        \mbox{For}     &   0 & -10 & -11 & -11 &-11 &-11 & -11 & -11 & -11 & \bm{-87}\\			
        \bottomrule
		\end{array}$$
		%\\\centerline{(b) 80th percentiles}
	\end{minipage}
	\caption{\label{t_CopComp3}Copeland's scores of the measure families for 80th percentiles}
\end{table}

One can observe that for different datasets, ranking of measure families w.r.t. the quality of clustering differs. In Table~\ref{t_CopComp2}, for six datasets, SCCT takes the 1st or 2nd place; logComm does so for five datasets. In most cases,
%four Newsgroup datasets, the leader is logComm, while
the logarithmic measures outperform the corresponding plain ones.
For the ``news\_3cl\_3'' dataset and 80th percentiles, the leaders are FE and logFor. %, whereas logComm is inferior to the other logarithmic measures and even to SP-CT, while SCCT is the worst one.
%On the contrary, ``news\_2cl\_3'' favors SCCT and then logComm.
For ``Zachary'' with the best parameter, all measures, except for For, reach an absolute result.
%For ``polbooks'', the leader is RSP followed by SP-CT and SCT.
For ``football'' (having 12 classes), %one can see a familiar picture in which
Comm and pWalk are the winners with the best parameters, like in the cases of two classes of different sizes (cf.\ Fig.\,\ref{f_difClas}(b)) and of six classes (Fig.\,\ref{f_6classes}); SCCT is the worst. \x{^}

The comparison of Tables~\ref{t_CopComp2} and \ref{t_CopComp3} demonstrates that the rather high results of logHeat and RSP are not stable enough, so in the ranking with 80th percentiles, they lose four and three positions, respectively; Walk, logFor and SP-CT shift up two places each.\x{^} This dynamics resembles that in Table~\ref{t_CopComp}(a), (b).

\section{Conclusion}
\label{s_Conclu}

The main conclusion of our study is that in most cases, %of our study,
including the simple cases of random graphs with homogeneous classes of similar size,
logarithmic measures (i.e., measures resulting after taking logarithm of the proximities) better reveal the underlying structure
%distinguish the underlying classes
than the ``plain'' measures do. A direct comparison of inter-class and intra-class distances by drawing the reject curves confirms this conclusion (with the exception of Heat and logHeat).\x{^}

In our experiments, the three leading measure families in the aforementioned simple cases, according to Copelands's test presented in Table~\ref{t_CopComp}, are logarithmic Communicability, Sigmoid Corrected Commute Time kernel, and logarithmic Forest\x{^}. %, logarithmic Forest, and Free Energy. The latter can also be considered as a kind of logarithmic measure: see the definition of $\Phi$ in Subsection~\ref{s_FE}.
The superiority of logarithmic Communicability over the other measures is observed here for all sets of random graphs, except for the set $(200, 2, 0.1),$ for which SCCT is the best.

A plausible explanation of the superiority of logarithmic measures is that most kernels and proximity measures under study have a multiplicative nature, while the nature of distances that cluster algorithms actually use\x{^} is an additive one (as the triangle inequality reveals). %suggests
The logarithmic transformation is precisely the tool that transforms the first nature to the second one. Moreover, some distances corresponding to the logarithmic measures possess a meaningful cutpoint additivity property.

At the same time, there are more complex\x{^} heterogeneous networks for which other measures can behave well. Among such structures, we can mention one type of networks with classes of different sizes and smaller classes of moderate sizes, for which two ``plain'' measures, Comm and pWalk can outperform the logarithmic measures under study. A similar situation can be observed for some structures with many classes.\x{^}
The SCCT kernel, which involves the sigmoid transformation instead of the logarithmic one, performs very well in many experiments.
In Ward's clustering (with the best parameter values) of several datasets, it even wins %outperforms logComm
in the competition by Copeland's score.
% however, as applied to several datasets, it is definitely inferior to logComm and the other logarithmic measures in reliability.

\bibliographystyle{spmpsci}%{alpha}%{plainnat}%{gost2008n}%{arc1}%{unsrt}%{SimpleU}
%{spphys} %without titles
%{abbrvnat} %OK
%alphaurl %alphabetical, without abbr.
%{alpha-letters} %gives big letters in the references
%{alphadin} % German
%{alphahtmldate} %- in HTML format
%{\scriptsize
\bibliography{all2}

%\begin{thebibliography}{99}
%\end{thebibliography}
\end{document}